\documentclass{article}
\usepackage{times}
\usepackage{multicol}
\usepackage[bookmarks=true]{hyperref}
\usepackage{graphicx}
\usepackage{float}
\usepackage{amssymb}
\usepackage[linesnumbered,ruled,vlined]{algorithm2e}
\usepackage{amsmath}
\usepackage{hyperref}
\usepackage{placeins}
\usepackage{caption}
\usepackage{subcaption} 
\usepackage{pifont}
\usepackage{diagbox}
\usepackage{wrapfig}
\usepackage[preprint]{corl_2025} % Uncomment for pre-prints (e.g., arxiv); This is like ``final'', but will remove the CORL footnote.

\title{GraspFactory: A Large Object-Centric \\ Grasping Dataset}

\author{
    Srinidhi Kalgundi Srinivas$^{\dagger}$,\hspace{1em}
    Yash Shukla$^{\text{\S}}$,\hspace{1em}
    Adam Arnold$^{\dagger}$,\hspace{1em}
    Sachin Chitta$^{\dagger}$ \\
    $^{\dagger}$Autodesk Research \\
    {\small{$^{\S}$Work done as an intern at Autodesk Research}}
}

\begin{document}
\maketitle

%===============================================================================

\begin{abstract}
    Robotic grasping is a crucial task in industrial automation, where robots are increasingly expected to handle a wide range of objects. However, a significant challenge arises when robot grasping models trained on limited datasets encounter novel objects. In real-world environments such as warehouses or manufacturing plants, the diversity of objects can be vast, and grasping models need to generalize to this diversity.  Training large, generalizable robot-grasping models requires geometrically diverse datasets. In this paper, we introduce GraspFactory, a dataset containing over 109 million 6-DoF grasps collectively for the Franka Panda (with 14,690 objects) and Robotiq 2F-85 grippers (with 33,710 objects). GraspFactory is designed for training data-intensive models, and we demonstrate the generalization capabilities of one such model trained on a subset of GraspFactory in both simulated and real-world settings. The dataset and tools are made available for download at \href{https://graspfactory.github.io/}{\textbf{\textcolor{blue}{graspfactory.github.io}}}.
\end{abstract}


%===============================================================================

\section{Introduction}
	
Large datasets have been a major contributor to the success of AI models. The fields of Computer Vision and Natural Language Processing have seen tremendous progress due to the presence of internet-scale datasets like ImageNet \cite{russakovsky2015imagenet} and Laion-5b \cite{schuhmann2022laion}. Models such as Chat-GPT \cite{openai2023chatgpt} and Dall-E\cite{openai2023dalle} demonstrate strong generalization capabilities for tasks that were not explicitly represented in their training data, thanks to the use of diverse training datasets and large-scale transformer-based architectures. Similar efforts have been undertaken in robotics to collect large datasets, such as Open X-Embodiment \cite{open_x_embodiment_rt_x_2023} and DROID \cite{khazatsky2024droid}. These datasets focus on end-to-end training of robots but there is still a need for task-specific datasets. Robot grasping is one such task, and a generalized grasping model remains elusive, in part due to the lack of geometrically diverse objects in existing datasets. In this work, we present an object-centric grasping dataset that offers greater geometric diversity compared to existing datasets.

Currently, object-centric grasping datasets \cite{eppner2021acronym, morrison2020egad, murali2025graspgen} and scene-based grasping datasets \cite{jiang2011cornell, depierre2018jacquard, fang2020graspnet1b} are mostly geared toward domestic robotics applications. These datasets have been used to train robot grasping models such as \cite{asif2018graspnet, urain2023se, sundermeyer2021contact, barad2024graspldm}. The grasping datasets are generated using 3D CAD models\footnote{We use the term ‘‘CAD models’’ in this paper to specifically refer to triangular meshes.} from 3D datasets such as Shapenet \cite{chang2015shapenet}, YCB \cite{calli2015benchmarking}, Objaverse \cite{deitke2023objaverse} and the Princeton Shape Benchmark \cite{shilane2004psb}. These datasets, however, contain objects of low geometric diversity, as they contain only a small number of semantic classes \cite{morrison2020egad}.  Some of the recent advancements in 3D generative models, however, are fueled by larger 3D datasets like those presented in \cite{Koch_2019_ABC, deitke2023objaverse, chang2015shapenet}. We leverage one such 3D dataset, \textit{ABC-Dataset} \cite{Koch_2019_ABC}, that contains \textit{1M+} high quality geometric models.

We introduce GraspFactory, a large-scale dataset of 6-DoF parallel-jaw grasps generated in simulation. The dataset provides two-fingered grasps for the Franka Panda and Robotiq 2F-85 grippers. We utilize a scalable robotics simulation and synthetic data generation tool to annotate the objects with 6-DoF grasps. Further, we train an existing diffusion-based grasping model, SE(3)-DiffusionFields \cite{urain2023se} on the Franka Panda subset, and evaluate the model's generalization capabilities on unseen objects. To the best of our knowledge, this is the largest object-centric grasping dataset containing 6-DoF, parallel-jaw grasps for geometrically diverse 3D data.

Our contributions are as follows:
\begin{itemize}
    \item We present GraspFactory, a large-scale, object-centric dataset of 6-DoF parallel-jaw grasps with corresponding gripper widths, comprising over 109 million grasps in total. The dataset includes grasps for 33,710 objects randomly selected from the ABC dataset \cite{Koch_2019_ABC} for the Robotiq 2F-85 gripper, and 14,690 objects for the Franka Panda gripper, selected as a subset of the Robotiq object set. As part of ongoing work, we plan to extend the dataset with grasps for additional objects from \cite{Koch_2019_ABC}.
    
    \item We train a diffusion-based grasp generative model \cite{urain2023se} on the Franka Panda subset of GraspFactory, and demonstrate that training on geometrically diverse data improves generalization in both simulation and real-world experiments.
\end{itemize}

The rest of the paper is organized as follows: In Section \ref{sec:related-work}, we review prior work. In Section \ref{sec:method}, we present the method used for generating the dataset. Section \ref{sec:results} describes the experimental setup, both in simulation and real, and the results from training a model with GraspFactory.

%===============================================================================
\section{Related Work}
\label{sec:related-work}
\subsection{\textbf{Existing datasets}}
Robot grasping datasets are generally collected through one of the following three methods: simulation, human annotation or human teleoperation. Datasets collected through simulation offer scalability, but they require highly accurate physics simulators to overcome the sim-to-real gap. Simulators built on physics engines such as PhysX \cite{physx} and Bullet \cite{coumans2015bullet} offer some level of physical realism. These simulators require CAD models of objects for scene generation. There are several datasets containing 3D CAD models \cite{Koch_2019_ABC, chang2015shapenet, deitke2023objaverse, willis2021fusion} that are currently used to train 3D generative models \cite{hui2024make, sanghi2024wavelet}, 3D segmentation and classification models \cite{qi2017pointnet}, and normal estimation methods \cite{Koch_2019_ABC}. 

Prior work \cite{eppner2021acronym, mahler2017dex, morrison2020egad} uses physics simulators and 3D datasets to generate grasping datasets. \citet{kappler2015leveraging} show that physics simulation can be used to predict successful grasps. \citet{eppner2021acronym} use ShapeNet \cite{chang2015shapenet}, \citet{mahler2017dex} use 3D-Net \cite{walter20123dnet} and the KIT object database \cite{alex2012kit} while \citet{murali2025graspgen} use the Objaverse dataset \cite{deitke2023objaverse} to generate grasping datasets. However, the objects in these datasets are primarily used for 3D object recognition tasks containing only a small number of semantic classes, resulting in low geometric diversity within the datasets \cite{morrison2020egad}. \citet{morrison2020egad} propose a method to use evolutionary algorithms to generate objects and grasps of varying complexities, but the generated objects are not representative of those encountered in the real-world. 

There are also several datasets in the literature that contain grasps for images and point clouds of scenes with multiple objects. \citet{jiang2011cornell} and \citet{depierre2018jacquard} propose datasets containing planar grasps in the image frame.  \citet{fang2020graspnet1b} present a dataset that contains over 1.1 billion grasp annotations for cluttered, complex scenes. \citet{zhang2019roi} expand the Visual Manipulation Relationship Dataset \cite{zhang2018vmrd} containing planar grasps for \(15k+\) objects. Despite the large number of grasp annotations in them, the planar nature of the grasps in these datasets limits their utility for tasks like bin-picking,  where the objects may not be presented on a plane. GraspFactory uses \cite{Koch_2019_ABC} to generate, to the best of our knowledge, the largest object-centric 6-DoF grasping dataset containing objects of varied geometries. 

\subsection{\textbf{Grasp Sampling Methods}}
Given an object CAD model, one  desired behavior of a grasp, is to maintain {\em force closure}  in the presence of disturbing forces and moments while respecting velocity and kinematic constraints of the manipulator \cite{Prattichizzo2008grasping}. A number of methods have been proposed in the literature to sample robust grasps from a CAD model. \citet{Gatrell1989CAD} uses the information from CAD models such as polygons, edges and vertices to generate grasps using Extended Gaussian Images \cite{horn1984egi} that achieve force closure. Other grasp sampling methods using CAD models include uniform samplers \cite{Lav06}, approach based samplers \cite{veres2017integrated}  and antipodal-based samplers \cite{pas2015using, mahler2017dex}. \citet{zhu2003synthesis} and \citet{han2000grasp} propose analytical approaches to test force-closure condition for the sampled grasps. \citet{eppner2021acronym} present a two-fingered grasp dataset using objects from \cite{chang2015shapenet}. They evaluate the sampled grasps using the FleX \cite{muller2014flex} physics simulator. Similar to \cite{eppner2021acronym}, we use a physics simulator to evaluate the robustness of sampled grasps under external wrenches. Our work uses the antipodal sampling method \cite{eppner2019billion} to sample 6-DoF parallel-jaw grasps, and uses the Isaac Sim simulator \cite{isaacsim2023} for evaluating grasp robustness.

\subsection{\textbf{Learning-based grasping}}
A number of deep learning-based methods have been proposed to estimate grasps from an object's CAD model and also directly from the scene point cloud. \citet{newbury2023deep} present a comprehensive survey on different methods and datasets used in the literature. Grasps are broadly classified into 4-DoF and 6-DoF, where 4-DoF planar grasp estimation methods involve determining the \textit{x, y, z and $\theta$} parameters, where, \textit{x, y, z} are the 3D spatial position and \textit{$\theta$} is the rotation about the z-axis of the gripper and 6-DoF grasp estimation methods involve determining the full 6D pose of the gripper for a suitable grasp.  \citet{morrison2018closing} proposed a convolution-based neural network for detecting suitable grasps from a depth image while also considering the width of the gripper as a parameter. \citet{mousavian20196} propose a Variational Auto-Encoder \cite{kingma2013auto} based model and sample grasps through the latent space of the model. \citet{sundermeyer2021contact} propose an end-to-end network to sample grasps from a depth image of the scene.  \citet{barad2024graspldm}, \citet{urain2023se}, and \citet{murali2025graspgen} use diffusion models to generate 6-DoF grasps.

%===============================================================================
\section{Approach: Generating the GraspFactory Dataset} \label{sec:method}
Our approach to the GraspFactory dataset generation involves grasp sampling to generate candidate grasps, collision checks to filter out grasps where the gripper may be in collision with the object and physics based evaluation to determine whether the grasp can hold the object firmly. This section provides a detailed description of our approach.  

\subsection{Object CAD Models}
\label{subsec:objects}
We source the CAD models used in this study from the ABC dataset, which contains 1M+ diverse objects. We choose to work with 33,710 randomly selected objects from the ABC dataset for the Robotiq 2F-85 gripper and a subset of 14,690 of these objects for the Franka Panda gripper.

\subsection{Grasp Sampling}
\label{subsec:grasp-sampling}

The first step in our approach is to sample candidate grasps for all the chosen objects. Given the CAD model for an object, we utilize the antipodal sampling method to sample grasps. We ensure that the CAD models are watertight using \cite{huang2018robust}. We sample points on the mesh surface and compute their surface normals, denoted as \textit{$\hat{n}$}. We then cast three rays within a cone aligned with the surface normal, with a vertex angle of $30^\circ$, and identify the points on the CAD model that these rays intersect. We only consider the points whose surface normal is in the opposite direction of the ray origin's surface normal, as these points represent potential antipodal contact points. The gripper pose is determined by aligning the fingers' surface normals with the line connecting these antipodal points, and the \textit{z}-axis of the gripper is aligned with four uniformly spaced vectors, each $90^\circ$ apart, around this line.

Additionally, we decimate the object CAD model by a factor of 0.6 and repeat the antipodal grasp sampling process described above. A mesh decimation factor of 0.6 enables us to preserve the underlying shape of the mesh while also increasing the number of potential graspable surfaces. We define a graspable surface on a mesh as a collection of triangles that come in contact with the finger grasping the object.

As shown in Fig. \ref{fig:collision-check}, we perform collision checks between the gripper in sampled grasp poses and the object's CAD model using an internally developed robotics research software platform, eliminating grasps where finger geometry collides with the object.

We sample a total of 391.38 million non-colliding grasp candidates across 33,710 objects from the ABC dataset \cite{Koch_2019_ABC} using this approach.

\subsection{Physics Based Grasp Evaluation}
\label{subsec:grasp-evaluation}

We evaluate the accuracy of each of the sampled grasps from Section \ref{subsec:grasp-sampling} in the Isaac Sim simulator \cite{isaacsim2023}. Due to computational limitations, we evaluate 2,000 grasps per object for Franka Panda robot equipped with a Franka hand as shown in Fig. \ref{fig:eval-grasp-exec} and 5,000 grasps per object for the Robotiq 2F-85 gripper as shown in Fig. \ref{fig:eval-isaac-sim-robotiq}. The evaluated grasps are selected using Agglomerative Hierarchical Clustering \cite{hang2014hierarchical} in the \( \text{SE}(3) \) space, which does not require a predefined number of clusters and can capture complex cluster structures. The distance \(d\) between two grasps \(g_1\) and \(g_2\) in the clustering process is defined as:

\begin{equation} \label{eq:distance-metric}
    d = \| \mathbf{t}_1 - \mathbf{t}_2 \| + \arccos\left(|q_1 \cdot q_2|\right)
\end{equation}

where \(\mathbf{t}_1\) and \(\mathbf{t}_2\) are the translation components of \(g_1\) and \(g_2\), respectively, in \(\mathbb{R}^3\). The orientations of \(g_1\) and \(g_2\) are represented as unit quaternions, \(q_1\) and \(q_2\) respectively.

We spawn 2,000 Franka Panda robots in the simulated environment, as shown Appendix \ref{Appendix:data-gen}, to test each of the selected grasps for 14,690 objects. For each grasp, we spawn the object such that the grasp's \(z\)-axis is aligned with the world \(z\)-axis.  We move the robot to the grasp pose and close the fingers around the object. Fig. \ref{fig:eval-grasp-exec} in Appendix \ref{Appendix:data-gen} shows an example of a successfully grasped object. To ensure that the grasp is robust against external forces, we move the robot through a set of pre-defined poses, effectively testing whether the grasp can withstand perturbations. We record some extra information from the simulation runs and include it in the dataset for possible future use. Using the contact force information from the simulated environment, we record grasps that are in contact with the fingers during the entire simulation, and also record contact forces exerted on each spawned objects. We also record the  duration of contact for the failed grasps. 

Additionally, we spawn 5,000 Robotiq 2F-85 grippers in the simulated environment, as shown in Fig. \ref{fig:eval-isaac-sim-robotiq} of Appendix \ref{Appendix:data-gen}, to test each of the selected grasps for 33,710 objects. We move the gripper along the positive and negative world \(z\)-axis and rotate the gripper about the world \(z\)-axis to test the grasp robustness against external forces. We record a grasp to be successful if the object remains in between the fingers at the end of the simulation.

Successfully evaluated grasps are also referred to as \textit{good} or \textit{feasible} grasps in this paper. 

Algorithm \ref{alg:grasp-generation} summarizes our approach to generating the GraspFactory dataset.

\begin{algorithm}[htbp] 
\DontPrintSemicolon
\begin{minipage}{0.9\columnwidth}
\caption{Grasp Sampling and Evaluation}
\label{alg:grasp-generation}
\SetKwInput{Input}{Input}
\SetKwInput{Output}{Output}
\SetKwFunction{Cluster}{Cluster}
\SetKwFunction{Evaluate}{Evaluate}

\Input{CAD Model \(M\), Decimation Factor \(d\)}
\Output{Successful Grasp Poses \{$_o$\(T\)$_g$\} with gripper widths \(\{g_w\}\) in mm}

\BlankLine
\textbf{Step 1: Grasp Sampling}\;
\ForEach{point \(p\) on \(M\)}{
    Compute surface normal \(\hat{n}\) at \(p\)\;
    Cast rays within \(30^\circ\) cone aligned with \(\hat{n}\)\;
    Find antipodal points \((p_1, p_2)\) satisfying \(\hat{n}_1 \cdot \hat{n}_2 = -1\)\;
    Align fingers with points \((p_1, p_2)\) and gripper \(z\)-axis spaced \(90^\circ\) apart\;
    \If{collision-free}{
        Add \((_oT_g, g_w)\)\ to list;
    }
}
Decimate \(M\) by factor \(d\) and repeat sampling\;

\BlankLine
\textbf{Step 2: Grasp Clustering}\;
Apply Agglomerative Hierarchical Clustering algorithm on sampled grasps in \( SE(3) \) space\;
Select \textit{N} representative grasps \(\{_oT_g\}\)\ and their corresponding gripper widths \(\{g_w\}\);

\BlankLine
\textbf{Step 3: Grasp Evaluation}\;
\ForEach{grasp \(_oT_g\) in \(\{_oT_g\}\)}{
    Spawn object in simulation such that \(z\)-axis of \(_oT_g\) is aligned with the world \(z\)-axis \;
    Move robot to \(_oT_g\) and close gripper\;
    Apply perturbations to test grasp robustness\;
    Record success and failures\;
}
\end{minipage}
\BlankLine
\textbf{Output:} Successful grasps with poses, failed grasps with poses and widths
\end{algorithm}

%===============================================================================

\subsection{Dataset Statistics} \label{subsec:dataset}
The GraspFactory dataset contains object-centric 6-DOF parallel grasps for the Franka Panda and Robotiq 2F-85 grippers. We include a list of grasp poses \textit{$_{o}$T$_{g}$} in the object coordinate frame and corresponding grasping width, and a list of indices of grasps that succeeded in physics simulation, \textit{$g_w$}, where: 

\begin{equation}
\label{eq:transformation_matrix}
\textit{$_{o}$T$_{g}$} = \left\{ 
\begin{bmatrix}
R & t \\
0 & 1
\end{bmatrix}
R \in \text{SO}(3), \; t \in \mathbb{R}^3 
\right\}
\end{equation}
Grasping width is defined as the distance between the fingers of the gripper when grasping an object, measured in \textit{mm}.

After physics based evaluation, GraspFactory contains 12.2 million feasible grasps for 14,690 objects for the Franka Panda gripper and 97.1 million feasible grasps for 33,710 objects for the Robotiq 2F-85 gripper, surpassing \cite{eppner2021acronym} in both the number of objects and grasps.

Given the size, diversity,  and the real-world nature of this dataset, it is well-suited for training grasping models. We present the results of the point cloud-based SE(3)-DiffusionFields model \cite{urain2023se} trained on Franka Panda subset of the GraspFactory dataset in Section \ref{sec:results}.

\subsection{Model Training} \label{subsec:model-training}
We train the point cloud-based model proposed in SE(3)-DiffusionFields using the Franka Panda Hand subset of GraspFactory and the ACRONYM datasets on two NVIDIA RTX-4090s with a batch size of 4 for 2,900 epochs over 19 days. We focus on the Franka Panda subset to align with the ACRONYM dataset, which also contains grasps for the Franka Panda gripper. SE(3)-Diffusion-Fields has been shown to outperform other grasp generative models in capturing and generating diverse grasps \cite{urain2023se}. We use the same learning rate scheduler provided by \cite{urain2023se}. We use 12,903 objects in the training set, 1,434 objects in the validation set, and 353 objects held out as part of the test set.

%===============================================================================
\section{Experiments and Results}
\label{sec:results}

We evaluate the model's performance on a set of industrial objects of varying geometric complexities that were not part of the training data. We compare the model trained using GraspFactory to the same model trained using the ACRONYM dataset. We test the performance in both simulation and real-world settings. Simulation allows testing a larger set of grasps, predicted by the model, for their accuracy, while real-world settings test the physical feasibility and robustness of the grasps.

\subsection{Simulated Experiments}

\begin{figure}[htbp]
    \begin{subfigure}[t]{0.48\columnwidth}
        \centering
        \includegraphics[height=3cm, keepaspectratio]{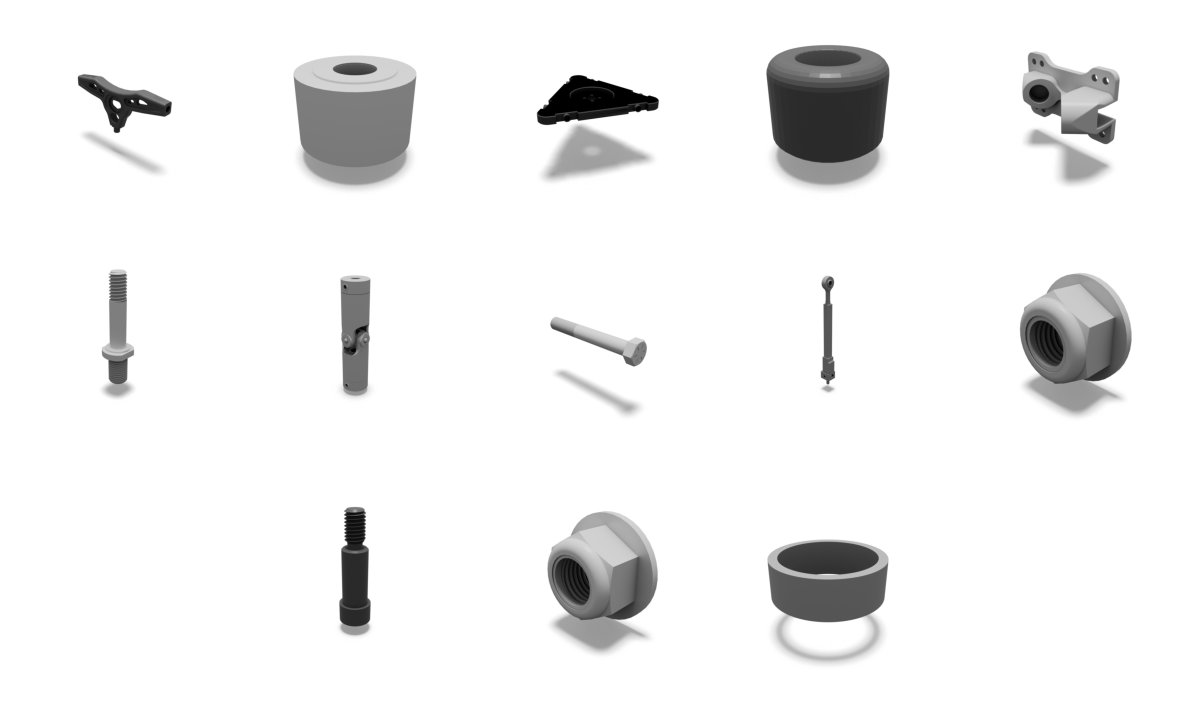}
        \captionsetup{justification=centering}
        \caption{\textbf{Top row}: Hanger, Hardcore Bearing, Top Plate, Wheel, Base\\
                 \textbf{Middle row}: Axle, Elbow Joint, Kingpin Bolt, Strut, Kingpin Nut\\
                 \textbf{Bottom row}: Shoulder Screw, Axle Nut, Hollow Cylinder}
        \label{fig:objects-in-sim}
    \end{subfigure}
    \hfill
    \begin{subfigure}[t]{0.48\columnwidth}
        \centering
        \includegraphics[height=3cm, keepaspectratio]{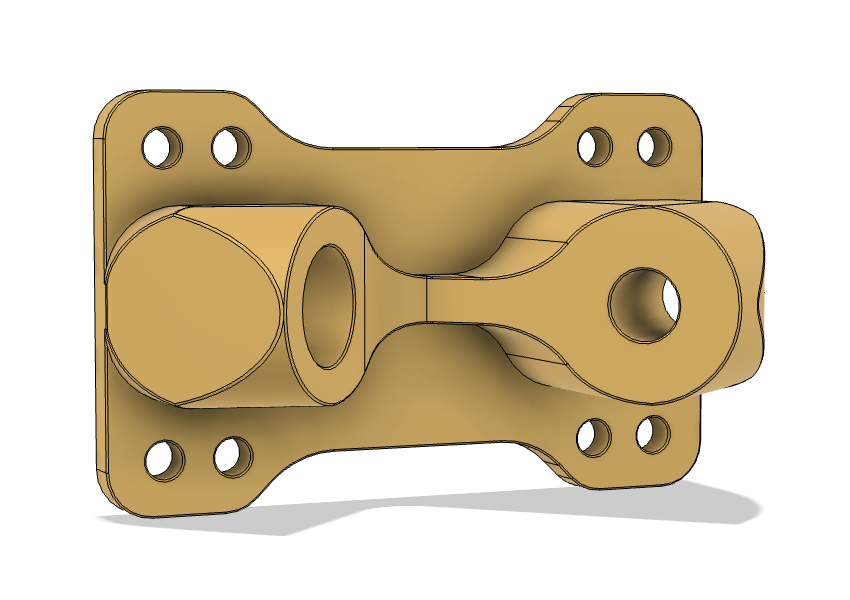}
        \captionsetup{justification=centering}
        \caption{Zoomed-in view of the Base, showing its complexity.}
        \label{fig:base}
    \end{subfigure}
    \caption{Examples of objects used in experiments and simulations.}
    \label{fig:combined-base-and-sim}
\end{figure}

We sample 100 grasps from the models trained on GraspFactory and ACRONYM for the objects shown in Fig. \ref{fig:objects-in-sim}. Since simulation is non-deterministic, there are minor differences in the absolute success rate numbers. To avoid any bias and to provide statistical consistency, we run the experiments with two random seeds. Qualitative evaluation shown in Appendix \ref{appendix:exp-and-results} demonstrates that the model trained on the ACRONYM dataset generates grasps that intersect with the object meshes, whereas the model trained on GraspFactory produces non-intersecting grasps, showing the effectiveness of our dataset for applications involving complex geometries.

Grasp success rate is evaluated using the same metrics to identify successful grasps as explained in Sec. \ref{subsec:grasp-evaluation} and define accuracy as the percentage of successful grasps in simulation. We show the success rate for 100 grasps sampled from the model trained on GraspFactory and ACRONYM datasets in Table \ref{tab:sim-results}. The model trained on GraspFactory outperforms the model trained on ACRONYM across all objects shown in Fig. \ref{fig:objects-in-sim} in simulation by a wide margin in most cases. In fact, the success rates are close only for objects like the \textit{Strut}, whose constituent geometric primitives (cuboid and cylinder) are well represented in the ACRONYM dataset.

\begin{table*}[htbp]
\centering
\renewcommand{\arraystretch}{0.9}
\begin{tabular}{c cc cc }
\hline
\textbf{Objects} & \multicolumn{2}{c }{\textbf{Round 1}} & \multicolumn{2}{c}{\textbf{Round 2}} \\
\hline
 & \textbf{GraspFactory (Ours)} & \textbf{ACRONYM} & \textbf{GraspFactory (Ours)} & \textbf{ACRONYM} \\
\hline
\hline
Hanger & 0.97 & 0.30 & 0.93  & 0.22 \\
\hline
Hardcore Bearing & 0.99  & 0.00 & 0.94 & 0.00 \\
\hline
Top Plate & 0.91 & 0.53 & 0.96 & 0.49 \\
\hline
Wheel & 1.00 & 0.00 & 1.00  & 0.00 \\
\hline
Base & 0.75 & 0.00 & 0.72  & 0.47 \\
\hline
Axle & 0.85 & 0.00 & 0.72  & 0.00 \\
\hline
Elbow Joint & 1.00  & 0.00 & 1.00 & 0.00 \\
\hline
Kingpin Bolt & 0.71 & 0.48 & 0.93  & 0.00 \\
\hline
Strut & 0.96 & 0.94 & 0.83  & 0.87 \\
\hline
Kingpin Nut & 0.84 & 0.00 & 0.85  & 0.00 \\
\hline
Shoulder Screw & 0.85 & 0.00  & 0.81 & 0.00 \\
\hline
Axle Nut & 0.99 & 0.00 & 0.96  & 0.00 \\
\hline
Hollow Cylinder & 0.97 & 0.40 & 0.91  & 0.55 \\
\hline
\end{tabular}
\captionsetup{justification=centering}
\caption{Simulation Results: Success rates for 100 grasps generated by the model trained on GraspFactory (Ours) and ACRONYM datasets in simulation.}
\label{tab:sim-results}
\end{table*}

The success rate in simulation is lower for \textit{Base}, shown in Fig. \ref{fig:base} compared to the other objects, as the flat finger geometry results in unstable grasps near its center of mass. Incorporating finger geometry may improve the model's ability to predict stable grasps for a specific finger geometry. \textit{Axle} and \textit{Kingpin Bolt} have hexagonal heads, and we observe that some of the grasp poses generated by the model are located over the vertices of the hexagon. In simulation, these grasps fail during grasp evaluation when we attempt to move the grasped object around (in the manner described in Section \ref{subsec:grasp-evaluation}).

\begin{table}[ht]
 \centering
 \renewcommand{\arraystretch}{0.8}
 {\small
    \begin{tabular}{l|cc}
        \hline
        \diagbox{Train Dataset}{Test Dataset} & GraspFactory & ACRONYM \\ \hline \hline
        GraspFactory (Ours)        & 0.75       & 0.59       \\ \hline
        ACRONYM           & 0.08       & 0.57       \\ \hline
    \end{tabular}
 }
\caption{Average success rates of 100 grasps for 353 and 95 test objects from GraspFactory (Ours) and ACRONYM respectively in simulation.}
\label{tab:eval}
\end{table}

Results of the model trained on GraspFactory and ACRONYM on held-out objects from both the datasets are in Table \ref{tab:eval}.

\subsection{Hardware Experiments} \label{sec:hardware-experiments}

We evaluate the strengths of GraspFactory using physical experiments based on two metrics: \textit{real-world feasibility} and \textit{grasp robustness} using the hardware setup and perception pipeline outlined in Appendix \ref{appendix:real-world-experiments}.\footnote{We use a UR10e robot and Robotiq 2F-85 gripper for testing on real-hardware due to an unanticipated lack of availability of our Franka Panda robot.}  We use \textit{real-world feasibility} to evaluate whether grasps generated by the model trained on GraspFactory can be used reliably to pick up an object without colliding with the object or the surrounding objects, such as the table. \textit{Grasp robustness} measures the consistency of a grasp across repeated trials, highlighting its ability to maintain performance under minor uncertainty introduced by perception.

For each object, we sample 200 grasps from the model. Using the perception pipeline described in Appendix \ref{appendix:real-world-experiments}, we obtain the pose of the object and grasps in the world-frame. Grasps that are in collision with the support surface (table) are then eliminated, resulting in a smaller non-colliding grasp set that we process further.

\begin{table*}[ht]
\centering
\resizebox{\textwidth}{!}{
\begin{tabular}{ c c c c c c c c c c}
\hline
\textbf{Parts} & \multicolumn{3}{c}{\textbf{Random Pose 1}} & \multicolumn{3}{c}{\textbf{Random Pose 2}} & \multicolumn{3}{c}{\textbf{Random Pose 3}} \\
\hline
 & \textbf{Num Eval} & \textbf{Num Success} & \textbf{Success Rate (\%)} 
 & \textbf{Num Eval} & \textbf{Num Success} & \textbf{Success Rate (\%)} 
 & \textbf{Num Eval} & \textbf{Num Success} & \textbf{Success Rate (\%)} \\
\hline
\hline
Strut & 30 & 29 & 96.67 & 30 & 29 & 96.67 & 30 & 30 & 100.00 \\ 
\hline
Elbow Joint & 30 & 28 & 93.33 & 30 & 30 & 100.00 & 30 & 29 & 96.67 \\ 
\hline
Wheel & 30 & 30 & 100.00 & 30 & 29 & 96.67 & 30 & 30 & 100.00 \\ 
\hline
Hanger & 30 & 30 & 100.00 & 30 & 30 & 100.00 & 30 & 30 & 100.00 \\ 
\hline
Base & 30 & 24 & 80.00 & 30 & 29 & 96.67 & 30 & 27 & 90.00 \\ 
\hline
Gear & 30 & 30 & 100.00 & 30 & 26 & 86.67 & 30 & 29 & 96.67 \\ 
\hline
Kingpin Bolt & 30 & 28 & 93.33 & 30 & 30 & 100.00 & 30 & 29 & 96.67 \\ 
\hline
Regrasp Fixture & 30 & 30 & 100.00 & 30 & 29 & 96.67 & 30 & 28 & 93.33 \\ 
\hline
\end{tabular}
}
\captionsetup{justification=centering}
\caption{Hardware results with real objects: Evaluation of real-world feasibility of grasps sampled from the model trained on GraspFactory for three random poses across eight parts.}
\label{tab:physical_accuracy}
\end{table*}

\begin{table*}[htbp]
\centering
\setlength{\tabcolsep}{10pt} 
\resizebox{\textwidth}{!}{
\begin{tabular}{ c c c c c c c}
\hline
\textbf{Part} & \textbf{Grasp 1} & \textbf{Grasp 2} & \textbf{Grasp 3} & \textbf{Grasp 4} & \textbf{Grasp 5} & \textbf{Average Success Rate (\%)} \\ 
\hline
\hline
Base & 10 & 10 & 10 & 8 & 10 & 96.00 \\ 
\hline
Hanger & 10 & 10 & 10 & 10 & 10 & 100.00 \\ 
\hline
Gear & 10 & 10 & 10 & 10 & 10 & 100.00 \\ 
\hline
Regrasp Fixture & 10 & 10 & 10 & 10 & 10 & 100.00 \\ 
\hline
\end{tabular}
}

\captionsetup{justification=centering}
\caption{Hardware results with real objects: Evaluation results for grasp robustness for five random non-colliding grasps. \textbf{Num Trials=10}}
\label{tab:grasp_robustness}
\end{table*}

\begin{table}[ht]
    \centering
     \renewcommand{\arraystretch}{0.8}
 {\small
    \begin{tabular}{c c}
        \hline
        \textbf{Part} & \textbf{Part} \\
        \hline \hline
        Axle (\textbf{5/5}) & Camera Mount B (\textbf{5/5}) \\
        GPU Cooling Bracket (\textbf{5/5}) & Drone RPM Sensor Mount (\textbf{5/5})  \\
        GPU Fan Bracket (\textbf{4/5}) & Drone Landing Gear Mount (\textbf{3/5}) \\
        Camera Mount A (\textbf{5/5}) & Drone Support Structure (\textbf{5/5}) \\
        Automotive Relay (\textbf{5/5}) & Drone Motor Mount (\textbf{5/5})  \\
        \hline
    \end{tabular}
    }
    \caption{Results of real world experiments (number of successful grasps / number of grasps tested).}
    \label{tab:real-world}
\end{table}

\paragraph{Real-World Feasibility} \label{para:phy_acc}
We evaluate real-world feasibility by randomly selecting 30 grasps per object from the non-colliding grasp set, resulting in a total of \(720\) grasps evaluated across eight objects for three random stable poses (shown in Fig. \ref{fig:real-world-experiment-poses} in Appendix \ref{appendix:real-world-experiments}). We consider a grasp to be successful if the gripper fingers do not collide with the object or the table, and the robot successfully picks up the object \(100 mm\) off the table and places it back. 

Table \ref{tab:physical_accuracy} shows the grasp success rate per object in each of the three selected poses.\footnote{Grasps generated by the ACRONYM-trained model were not evaluated, since qualitative inspection showed poor grasp quality.} We show that the model trained on GraspFactory produces grasps that can be executed in the real-world, demonstrating the real-world usefulness of the dataset.

We run our perception pipeline in an open-loop manner, meaning that we do not estimate the pose of the object when it is placed back on the support surface. We observe a minor change in the pose of the object (due to its shape) when the robot places the object back, resulting in subsequent picks to fail sometimes (without human intervention to restore the object to its original location). This is particularly pronounced for \textit{Base} in \textit{Pose 1} and \textit{Gear} in \textit{Pose 2} as shown in Table \ref{tab:physical_accuracy}. Tested poses of objects are shown in Appendix \ref{appendix:real-world-experiments}.

\paragraph{Grasp Robustness}
Grasp robustness is evaluated by selecting five grasps from the non-colliding grasp set for four objects. We perform 10 trials per grasp, where each trial involves picking up the object and placing it down, resulting in a total of \(200\) grasp evaluations. We use the same metrics as outlined in Section \ref{para:phy_acc} for grasp success and note the average success rate across the five selected grasps in Table \ref{tab:grasp_robustness}. Our results show that the tested grasps are fairly robust even with small perturbations in object pose. 

We also evaluate five grasps per each of the 10 additional objects of varying geometric complexities shown in Appendix \ref{appendix:real-world-experiments}'s Fig. \ref{fig:real-objects} and present our results in Table \ref{tab:real-world}.

One of the challenges and limitations in our real-world experiments is differentiating errors due to pose estimation from those caused by grasp estimation. Any calibration error also affects our estimates of where the objects are, hence affecting the success of our chosen grasps as well. We note, however, that pose estimation is not the focus of our work presented here.

%===============================================================================

%===============================================================================
\section{Conclusion}
\label{sec:conclusion}

In this paper, we introduce GraspFactory, a large parallel-jaw grasp dataset containing 
\(12.2\) million feasible grasps for the Franka Panda gripper across 14,690 geometrically diverse objects and \(97.1\) million feasible grasps for the Robotiq 2F-85 gripper across 33,710 objects. The geometric diversity of the dataset addresses a critical gap in existing grasp datasets, which focus on objects with limited shape complexity or variety. Our results in simulation show that a model trained on GraspFactory significantly outperforms a model trained on existing datasets, such as ACRONYM, in terms of generalization to novel objects. Furthermore, we evaluated more than \(900\) grasps generated by the model trained on GraspFactory in real-world settings, demonstrating that our dataset enables models to generate grasps that can be used reliably in the real-world. GraspFactory contains information about grasping width for each grasp pose, which could be used to learn collision-free grasps in cluttered scenarios.

In the future, we plan to integrate finger geometry into training, enhancing both feasibility and robustness of generated grasps. We also aim to extend our dataset to include a larger number of objects from the ABC dataset. Subsequently, we aim to extend GraspFactory to include grasps for other end-effectors, such as suction-cup grippers. This extension would enhance the dataset's versatility, enabling researchers to develop and evaluate grasping algorithms applicable to a wider range of robotic systems and applications. 
%===============================================================================

\bibliography{example}  % .bib

\begin{thebibliography}{56}
\providecommand{\natexlab}[1]{#1}
\providecommand{\url}[1]{\texttt{#1}}
\expandafter\ifx\csname urlstyle\endcsname\relax
  \providecommand{\doi}[1]{doi: #1}\else
  \providecommand{\doi}{doi: \begingroup \urlstyle{rm}\Url}\fi

\bibitem[Russakovsky et~al.(2015)Russakovsky, Deng, Su, Krause, Satheesh, Ma, Huang, Karpathy, Khosla, Bernstein, et~al.]{russakovsky2015imagenet}
O.~Russakovsky, J.~Deng, H.~Su, J.~Krause, S.~Satheesh, S.~Ma, Z.~Huang, A.~Karpathy, A.~Khosla, M.~Bernstein, et~al.
\newblock \href{https://arxiv.org/abs/1409.0575}{Imagenet large scale visual recognition challenge}.
\newblock \emph{International journal of computer vision}, 115:\penalty0 211--252, 2015.

\bibitem[Schuhmann et~al.(2022)Schuhmann, Beaumont, Vencu, Gordon, Wightman, Cherti, Coombes, Katta, Mullis, Wortsman, et~al.]{schuhmann2022laion}
C.~Schuhmann, R.~Beaumont, R.~Vencu, C.~Gordon, R.~Wightman, M.~Cherti, T.~Coombes, A.~Katta, C.~Mullis, M.~Wortsman, et~al.
\newblock \href{https://laion.ai/blog/laion-5b/}{Laion-5b: An open large-scale dataset for training next generation image-text models}.
\newblock \emph{Advances in Neural Information Processing Systems}, 35:\penalty0 25278--25294, 2022.

\bibitem[{OpenAI}(2023{\natexlab{a}})]{openai2023chatgpt}
{OpenAI}.
\newblock Chatgpt: Large language model.
\newblock \url{https://chat.openai.com/}, 2023{\natexlab{a}}.

\bibitem[{OpenAI}(2023{\natexlab{b}})]{openai2023dalle}
{OpenAI}.
\newblock Dall·e: Image generation model.
\newblock \url{https://openai.com/dall-e}, 2023{\natexlab{b}}.

\bibitem[Collaboration et~al.(2023)Collaboration, O'Neill, Rehman, Gupta, Maddukuri, Gupta, Padalkar, Lee, Pooley, and \textit{et al.}]{open_x_embodiment_rt_x_2023}
O.~X.-E. Collaboration, A.~O'Neill, A.~Rehman, A.~Gupta, A.~Maddukuri, A.~Gupta, A.~Padalkar, A.~Lee, A.~Pooley, and \textit{et al.}
\newblock \href{https://arxiv.org/abs/2310.08864}{Open {X-E}mbodiment: Robotic Learning Datasets and {RT-X} Models}.
\newblock \url{https://arxiv.org/abs/2310.08864}, 2023.

\bibitem[Khazatsky et~al.(2024)Khazatsky, Pertsch, Nair, Balakrishna, Dasari, Karamcheti, Nasiriany, Srirama, Chen, Ellis, and \textit{et al.}]{khazatsky2024droid}
A.~Khazatsky, K.~Pertsch, S.~Nair, A.~Balakrishna, S.~Dasari, S.~Karamcheti, S.~Nasiriany, M.~K. Srirama, L.~Y. Chen, K.~Ellis, and \textit{et al.}
\newblock \href{https://arxiv.org/abs/2403.12945}{Droid: A large-scale in-the-wild robot manipulation dataset}.
\newblock \emph{arXiv preprint arXiv:2403.12945}, 2024.

\bibitem[Eppner et~al.(2021)Eppner, Mousavian, and Fox]{eppner2021acronym}
C.~Eppner, A.~Mousavian, and D.~Fox.
\newblock \href{https://arxiv.org/abs/2011.09584}{Acronym: A large-scale grasp dataset based on simulation}.
\newblock In \emph{2021 IEEE International Conference on Robotics and Automation (ICRA)}, pages 6222--6227. IEEE, 2021.

\bibitem[Morrison et~al.(2020)Morrison, Corke, and Leitner]{morrison2020egad}
D.~Morrison, P.~Corke, and J.~Leitner.
\newblock \href{https://arxiv.org/abs/2003.01314}{Egad! an evolved grasping analysis dataset for diversity and reproducibility in robotic manipulation}.
\newblock \emph{IEEE Robotics and Automation Letters}, 5\penalty0 (3):\penalty0 4368--4375, 2020.

\bibitem[Murali et~al.(2025)Murali, Sundaralingam, Chao, Yamada, Yuan, Carlson, Ramos, Birchfield, Fox, and Eppner]{murali2025graspgen}
A.~Murali, B.~Sundaralingam, Y.-W. Chao, J.~Yamada, W.~Yuan, M.~Carlson, F.~Ramos, S.~Birchfield, D.~Fox, and C.~Eppner.
\newblock Graspgen: A diffusion-based framework for 6-dof grasping with on-generator training.
\newblock \emph{arXiv preprint arXiv:2507.13097}, 2025.
\newblock URL \url{https://arxiv.org/abs/2507.13097}.

\bibitem[Jiang et~al.(2011)Jiang, Moseson, and Saxena]{jiang2011cornell}
Y.~Jiang, S.~Moseson, and A.~Saxena.
\newblock \href{https://ieeexplore.ieee.org/document/5980145}{Efficient grasping from RGBD images: Learning using a new rectangle representation}.
\newblock In \emph{2011 IEEE International Conference on Robotics and Automation}, pages 3304--3311, 2011.
\newblock \doi{10.1109/ICRA.2011.5980145}.

\bibitem[Depierre et~al.(2018)Depierre, Dellandr{\'e}a, and Chen]{depierre2018jacquard}
A.~Depierre, E.~Dellandr{\'e}a, and L.~Chen.
\newblock \href{https://arxiv.org/abs/1803.11469}{Jacquard: A large scale dataset for robotic grasp detection}.
\newblock In \emph{2018 IEEE/RSJ International Conference on Intelligent Robots and Systems (IROS)}, pages 3511--3516. IEEE, 2018.

\bibitem[Fang et~al.(2020)Fang, Wang, Gou, and Lu]{fang2020graspnet1b}
H.-S. Fang, C.~Wang, M.~Gou, and C.~Lu.
\newblock \href{https://openaccess.thecvf.com/content_CVPR_2020/papers/Fang_GraspNet-1Billion_A_Large-Scale_Benchmark_for_General_Object_Grasping_CVPR_2020_paper.pdf}{GraspNet-1Billion: A Large-Scale Benchmark for General Object Grasping}.
\newblock In \emph{Proceedings of the IEEE/CVF Conference on Computer Vision and Pattern Recognition}, pages 11444--11453, 2020.

\bibitem[Asif et~al.(2018)Asif, Tang, and Harrer]{asif2018graspnet}
U.~Asif, J.~Tang, and S.~Harrer.
\newblock \href{https://www.ijcai.org/proceedings/2018/0677.pdf}{GraspNet: An Efficient Convolutional Neural Network for Real-time Grasp Detection for Low-powered Devices.}
\newblock In \emph{IJCAI}, volume~7, pages 4875--4882, 2018.

\bibitem[Urain et~al.(2023)Urain, Funk, Peters, and Chalvatzaki]{urain2023se}
J.~Urain, N.~Funk, J.~Peters, and G.~Chalvatzaki.
\newblock \href{https://arxiv.org/abs/2209.03855}{Se (3)-diffusionfields: Learning smooth cost functions for joint grasp and motion optimization through diffusion}.
\newblock In \emph{2023 IEEE International Conference on Robotics and Automation (ICRA)}, pages 5923--5930. IEEE, 2023.

\bibitem[Sundermeyer et~al.(2021)Sundermeyer, Mousavian, Triebel, and Fox]{sundermeyer2021contact}
M.~Sundermeyer, A.~Mousavian, R.~Triebel, and D.~Fox.
\newblock \href{https://arxiv.org/abs/2103.14127}{Contact-graspnet: Efficient 6-dof grasp generation in cluttered scenes}.
\newblock In \emph{2021 IEEE International Conference on Robotics and Automation (ICRA)}, pages 13438--13444. IEEE, 2021.

\bibitem[Barad et~al.(2024)Barad, Orsula, Richard, Dentler, Olivares-Mendez, and Martinez]{barad2024graspldm}
K.~R. Barad, A.~Orsula, A.~Richard, J.~Dentler, M.~Olivares-Mendez, and C.~Martinez.
\newblock \href{https://arxiv.org/abs/2312.11243}{Graspldm: Generative 6-dof grasp synthesis using latent diffusion models}.
\newblock \emph{IEEE Access}, 2024.

\bibitem[Chang et~al.(2015)Chang, Funkhouser, Guibas, Hanrahan, Huang, Li, Savarese, Savva, Song, Su, et~al.]{chang2015shapenet}
A.~X. Chang, T.~Funkhouser, L.~Guibas, P.~Hanrahan, Q.~Huang, Z.~Li, S.~Savarese, M.~Savva, S.~Song, H.~Su, et~al.
\newblock \href{https://arxiv.org/abs/1512.03012}{Shapenet: An information-rich 3d model repository}.
\newblock \emph{arXiv preprint arXiv:1512.03012}, 2015.

\bibitem[Calli et~al.(2015)Calli, Walsman, Singh, Srinivasa, Abbeel, and Dollar]{calli2015benchmarking}
B.~Calli, A.~Walsman, A.~Singh, S.~Srinivasa, P.~Abbeel, and A.~M. Dollar.
\newblock Benchmarking in manipulation research: The ycb object and model set and benchmarking protocols.
\newblock \emph{arXiv preprint arXiv:1502.03143}, 2015.

\bibitem[Deitke et~al.(2023)Deitke, Schwenk, Salvador, Weihs, Michel, VanderBilt, Schmidt, Ehsani, Kembhavi, and Farhadi]{deitke2023objaverse}
M.~Deitke, D.~Schwenk, J.~Salvador, L.~Weihs, O.~Michel, E.~VanderBilt, L.~Schmidt, K.~Ehsani, A.~Kembhavi, and A.~Farhadi.
\newblock \href{https://arxiv.org/abs/2212.08051}{Objaverse: A universe of annotated 3d objects}.
\newblock In \emph{Proceedings of the IEEE/CVF Conference on Computer Vision and Pattern Recognition}, pages 13142--13153, 2023.

\bibitem[Shilane et~al.(2004)Shilane, Min, Kazhdan, and Funkhouser]{shilane2004psb}
P.~Shilane, P.~Min, M.~Kazhdan, and T.~Funkhouser.
\newblock \href{https://shape.cs.princeton.edu/benchmark/benchmark.pdf}{The Princeton Shape Benchmark}.
\newblock In \emph{Proceedings Shape Modeling Applications, 2004.}, pages 167--178, 2004.
\newblock \doi{10.1109/SMI.2004.1314504}.

\bibitem[Koch et~al.(2019)Koch, Matveev, Jiang, Williams, Artemov, Burnaev, Alexa, Zorin, and Panozzo]{Koch_2019_ABC}
S.~Koch, A.~Matveev, Z.~Jiang, F.~Williams, A.~Artemov, E.~Burnaev, M.~Alexa, D.~Zorin, and D.~Panozzo.
\newblock \href{https://arxiv.org/abs/1812.06216}{ABC: A Big CAD Model Dataset For Geometric Deep Learning}.
\newblock In \emph{The IEEE Conference on Computer Vision and Pattern Recognition (CVPR)}, June 2019.

\bibitem[{NVIDIA Corporation}(2019)]{physx}
{NVIDIA Corporation}.
\newblock \href{https://developer.nvidia.com/physx-sdk}{PhysX SDK 4.1}, 2019.
\newblock URL \url{https://developer.nvidia.com/physx-sdk}.
\newblock Accessed: 2025-01-06.

\bibitem[Coumans(2015)]{coumans2015bullet}
E.~Coumans.
\newblock \href{https://dl.acm.org/doi/10.1145/2776880.2792704}{Bullet physics simulation}.
\newblock In \emph{Special Interest Group on Computer Graphics and Interactive Techniques Conference, {SIGGRAPH} '15, Los Angeles, CA, USA, August 9-13, 2015, Courses}, page 7:1. {ACM}, 2015.
\newblock \doi{10.1145/2776880.2792704}.
\newblock URL \url{https://doi.org/10.1145/2776880.2792704}.

\bibitem[Willis et~al.(2021)Willis, Pu, Luo, Chu, Du, Lambourne, Solar-Lezama, and Matusik]{willis2021fusion}
K.~D. Willis, Y.~Pu, J.~Luo, H.~Chu, T.~Du, J.~G. Lambourne, A.~Solar-Lezama, and W.~Matusik.
\newblock \href{https://arxiv.org/abs/2010.02392}{Fusion 360 gallery: A dataset and environment for programmatic cad construction from human design sequences}.
\newblock \emph{ACM Transactions on Graphics (TOG)}, 40\penalty0 (4):\penalty0 1--24, 2021.

\bibitem[Hui et~al.(2024)Hui, Sanghi, Rampini, Malekshan, Liu, Shayani, and Fu]{hui2024make}
K.-H. Hui, A.~Sanghi, A.~Rampini, K.~R. Malekshan, Z.~Liu, H.~Shayani, and C.-W. Fu.
\newblock \href{https://arxiv.org/abs/2401.11067}{Make-a-shape: a ten-million-scale 3d shape model}.
\newblock In \emph{Forty-first International Conference on Machine Learning}, 2024.

\bibitem[Sanghi et~al.(2024)Sanghi, Khani, Reddy, Rampini, Cheung, Malekshan, Madan, and Shayani]{sanghi2024wavelet}
A.~Sanghi, A.~Khani, P.~Reddy, A.~Rampini, D.~Cheung, K.~R. Malekshan, K.~Madan, and H.~Shayani.
\newblock \href{https://arxiv.org/abs/2411.08017}{Wavelet Latent Diffusion (Wala): Billion-Parameter 3D Generative Model with Compact Wavelet Encodings}.
\newblock \emph{arXiv preprint arXiv:2411.08017}, 2024.

\bibitem[Qi et~al.(2017)Qi, Su, Mo, and Guibas]{qi2017pointnet}
C.~R. Qi, H.~Su, K.~Mo, and L.~J. Guibas.
\newblock \href{https://arxiv.org/abs/1612.00593}{Pointnet: Deep learning on point sets for 3d classification and segmentation}.
\newblock In \emph{Proceedings of the IEEE conference on computer vision and pattern recognition}, pages 652--660, 2017.

\bibitem[Mahler et~al.(2017)Mahler, Liang, Niyaz, Laskey, Doan, Liu, Ojea, and Goldberg]{mahler2017dex}
J.~Mahler, J.~Liang, S.~Niyaz, M.~Laskey, R.~Doan, X.~Liu, J.~A. Ojea, and K.~Goldberg.
\newblock \href{https://arxiv.org/abs/1703.09312}{Dex-net 2.0: Deep learning to plan robust grasps with synthetic point clouds and analytic grasp metrics}.
\newblock \emph{arXiv preprint arXiv:1703.09312}, 2017.

\bibitem[Kappler et~al.(2015)Kappler, Bohg, and Schaal]{kappler2015leveraging}
D.~Kappler, J.~Bohg, and S.~Schaal.
\newblock \href{https://ieeexplore.ieee.org/document/7139793}{Leveraging big data for grasp planning}.
\newblock In \emph{2015 IEEE International Conference on Robotics and Automation (ICRA)}, pages 4304--4311, 2015.
\newblock \doi{10.1109/ICRA.2015.7139793}.

\bibitem[Wohlkinger et~al.(2012)Wohlkinger, Aldoma, Rusu, and Vincze]{walter20123dnet}
W.~Wohlkinger, A.~Aldoma, R.~B. Rusu, and M.~Vincze.
\newblock \href{https://ieeexplore.ieee.org/document/6225116}{3DNet: Large-scale object class recognition from CAD models}.
\newblock In \emph{2012 IEEE International Conference on Robotics and Automation}, pages 5384--5391, 2012.
\newblock \doi{10.1109/ICRA.2012.6225116}.

\bibitem[Kasper et~al.(2012)Kasper, Xue, and Dillmann]{alex2012kit}
A.~Kasper, Z.~Xue, and R.~Dillmann.
\newblock \href{https://journals.sagepub.com/doi/10.1177/0278364912445831}{The KIT object models database: An object model database for object recognition, localization and manipulation in service robotics}.
\newblock \emph{The International Journal of Robotics Research}, 31\penalty0 (8):\penalty0 927--934, 2012.
\newblock \doi{10.1177/0278364912445831}.
\newblock URL \url{https://doi.org/10.1177/0278364912445831}.

\bibitem[Zhang et~al.(2019)Zhang, Lan, Bai, Zhou, Tian, and Zheng]{zhang2019roi}
H.~Zhang, X.~Lan, S.~Bai, X.~Zhou, Z.~Tian, and N.~Zheng.
\newblock \href{https://arxiv.org/abs/1808.10313}{Roi-based robotic grasp detection for object overlapping scenes}.
\newblock In \emph{2019 IEEE/RSJ International Conference on Intelligent Robots and Systems (IROS)}, pages 4768--4775. IEEE, 2019.

\bibitem[Zhang et~al.(2018)Zhang, Lan, Zhou, Tian, Zhang, and Zheng]{zhang2018vmrd}
H.~Zhang, X.~Lan, X.~Zhou, Z.~Tian, Y.~Zhang, and N.~Zheng.
\newblock \href{https://ieeexplore.ieee.org/document/8625071}{Visual Manipulation Relationship Network for Autonomous Robotics}.
\newblock In \emph{2018 IEEE-RAS 18th International Conference on Humanoid Robots (Humanoids)}, pages 118--125, 2018.
\newblock \doi{10.1109/HUMANOIDS.2018.8625071}.

\bibitem[Prattichizzo and Trinkle(2008)]{Prattichizzo2008grasping}
D.~Prattichizzo and J.~C. Trinkle.
\newblock \emph{\href{https://link.springer.com/referenceworkentry/10.1007/978-3-540-30301-5_29}{Grasping}}, pages 671--700.
\newblock Springer Berlin Heidelberg, Berlin, Heidelberg, 2008.
\newblock ISBN 978-3-540-30301-5.
\newblock \doi{10.1007/978-3-540-30301-5_29}.
\newblock URL \url{https://doi.org/10.1007/978-3-540-30301-5_29}.

\bibitem[Gatrell(1989)]{Gatrell1989CAD}
L.~Gatrell.
\newblock \href{https://ieeexplore.ieee.org/document/99987/}{CAD-based grasp synthesis utilizing polygons, edges and vertexes}.
\newblock In \emph{Proceedings, 1989 International Conference on Robotics and Automation}, pages 184--189 vol.1, 1989.
\newblock \doi{10.1109/ROBOT.1989.99987}.

\bibitem[Horn(1984)]{horn1984egi}
B.~Horn.
\newblock \href{https://ieeexplore.ieee.org/document/1457341}{Extended Gaussian images}.
\newblock \emph{Proceedings of the IEEE}, 72\penalty0 (12):\penalty0 1671--1686, 1984.
\newblock \doi{10.1109/PROC.1984.13073}.

\bibitem[LaValle(2006)]{Lav06}
S.~M. LaValle.
\newblock \emph{\href{https://msl.cs.uiuc.edu/planning/bookbig.pdf}{Planning Algorithms}}.
\newblock Cambridge University Press, Cambridge, U.K., 2006.
\newblock Available at http://planning.cs.uiuc.edu/.

\bibitem[Veres et~al.(2017)Veres, Moussa, and Taylor]{veres2017integrated}
M.~Veres, M.~Moussa, and G.~W. Taylor.
\newblock \href{https://arxiv.org/abs/1702.02103}{An integrated simulator and dataset that combines grasping and vision for deep learning}.
\newblock \emph{arXiv preprint arXiv:1702.02103}, 2017.

\bibitem[Pas and Platt(2015)]{pas2015using}
A.~t. Pas and R.~Platt.
\newblock \href{https://arxiv.org/abs/1501.03100}{Using geometry to detect grasps in 3d point clouds}.
\newblock \emph{arXiv preprint arXiv:1501.03100}, 2015.

\bibitem[Zhu and Wang(2003)]{zhu2003synthesis}
X.~Zhu and J.~Wang.
\newblock \href{https://ieeexplore.ieee.org/document/1220716}{Synthesis of force-closure grasps on 3-D objects based on the Q distance}.
\newblock \emph{IEEE Transactions on robotics and Automation}, 19\penalty0 (4):\penalty0 669--679, 2003.

\bibitem[Han et~al.(2000)Han, Trinkle, and Li]{han2000grasp}
L.~Han, J.~C. Trinkle, and Z.~X. Li.
\newblock \href{https://ieeexplore.ieee.org/document/897778}{Grasp analysis as linear matrix inequality problems}.
\newblock \emph{IEEE Transactions on Robotics and Automation}, 16\penalty0 (6):\penalty0 663--674, 2000.

\bibitem[Macklin et~al.(2014)Macklin, M\"{u}ller, Chentanez, and Kim]{muller2014flex}
M.~Macklin, M.~M\"{u}ller, N.~Chentanez, and T.-Y. Kim.
\newblock \href{https://dl.acm.org/doi/10.1145/2601097.2601152}{Unified particle physics for real-time applications}.
\newblock \emph{ACM Trans. Graph.}, 33\penalty0 (4), July 2014.
\newblock ISSN 0730-0301.
\newblock \doi{10.1145/2601097.2601152}.
\newblock URL \url{https://doi.org/10.1145/2601097.2601152}.

\bibitem[Eppner et~al.(2019)Eppner, Mousavian, and Fox]{eppner2019billion}
C.~Eppner, A.~Mousavian, and D.~Fox.
\newblock \href{https://arxiv.org/abs/1912.05604}{A billion ways to grasp: An evaluation of grasp sampling schemes on a dense, physics-based grasp data set}.
\newblock In \emph{The International Symposium of Robotics Research}, pages 890--905. Springer, 2019.

\bibitem[Corporation(2023)]{isaacsim2023}
N.~Corporation.
\newblock \emph{\href{https://developer.nvidia.com/isaac/sim}{NVIDIA Isaac Sim}}, 2023.
\newblock URL \url{https://developer.nvidia.com/isaac-sim}.
\newblock Version 2023.1.

\bibitem[Newbury et~al.(2023)Newbury, Gu, Chumbley, Mousavian, Eppner, Leitner, Bohg, Morales, Asfour, Kragic, et~al.]{newbury2023deep}
R.~Newbury, M.~Gu, L.~Chumbley, A.~Mousavian, C.~Eppner, J.~Leitner, J.~Bohg, A.~Morales, T.~Asfour, D.~Kragic, et~al.
\newblock \href{https://arxiv.org/abs/2207.02556}{Deep learning approaches to grasp synthesis: A review}.
\newblock \emph{IEEE Transactions on Robotics}, 39\penalty0 (5):\penalty0 3994--4015, 2023.

\bibitem[Morrison et~al.(2018)Morrison, Corke, and Leitner]{morrison2018closing}
D.~Morrison, P.~Corke, and J.~Leitner.
\newblock \href{https://arxiv.org/abs/1804.05172}{Closing the loop for robotic grasping: A real-time, generative grasp synthesis approach}.
\newblock \emph{arXiv preprint arXiv:1804.05172}, 2018.

\bibitem[Mousavian et~al.(2019)Mousavian, Eppner, and Fox]{mousavian20196}
A.~Mousavian, C.~Eppner, and D.~Fox.
\newblock \href{https://arxiv.org/abs/1905.10520}{6-dof graspnet: Variational grasp generation for object manipulation}.
\newblock In \emph{Proceedings of the IEEE/CVF international conference on computer vision}, pages 2901--2910, 2019.

\bibitem[Kingma(2013)]{kingma2013auto}
D.~P. Kingma.
\newblock \href{https://arxiv.org/abs/1312.6114}{Auto-encoding variational bayes}.
\newblock \emph{arXiv preprint arXiv:1312.6114}, 2013.

\bibitem[Huang et~al.(2018)Huang, Su, and Guibas]{huang2018robust}
J.~Huang, H.~Su, and L.~Guibas.
\newblock \href{https://arxiv.org/abs/1802.01698}{Robust watertight manifold surface generation method for shapenet models}.
\newblock \emph{arXiv preprint arXiv:1802.01698}, 2018.

\bibitem[Hang et~al.(2014)Hang, Stork, and Kragic]{hang2014hierarchical}
K.~Hang, J.~A. Stork, and D.~Kragic.
\newblock \href{https://ieeexplore.ieee.org/document/6942775}{Hierarchical Fingertip Space for multi-fingered precision grasping}.
\newblock In \emph{2014 IEEE/RSJ International Conference on Intelligent Robots and Systems}, pages 1641--1648, 2014.
\newblock \doi{10.1109/IROS.2014.6942775}.

\bibitem[Goldfeder et~al.(2009)Goldfeder, Ciocarlie, Dang, and Allen]{goldfeder2009columbia}
C.~Goldfeder, M.~Ciocarlie, H.~Dang, and P.~K. Allen.
\newblock \href{https://ieeexplore.ieee.org/document/5152709}{The Columbia grasp database}.
\newblock In \emph{2009 IEEE International Conference on Robotics and Automation}, pages 1710--1716, 2009.
\newblock \doi{10.1109/ROBOT.2009.5152709}.

\bibitem[Nguyen et~al.(2023)Nguyen, Groueix, Ponimatkin, Lepetit, and Hodan]{nguyen2023cnos}
V.~N. Nguyen, T.~Groueix, G.~Ponimatkin, V.~Lepetit, and T.~Hodan.
\newblock \href{https://arxiv.org/abs/2307.11067}{Cnos: A strong baseline for cad-based novel object segmentation}.
\newblock In \emph{Proceedings of the IEEE/CVF International Conference on Computer Vision}, pages 2134--2140, 2023.

\bibitem[Kirillov et~al.(2023)Kirillov, Mintun, Ravi, Mao, Rolland, Gustafson, Xiao, Whitehead, Berg, Lo, Doll{\'a}r, and Girshick]{kirillov2023segany}
A.~Kirillov, E.~Mintun, N.~Ravi, H.~Mao, C.~Rolland, L.~Gustafson, T.~Xiao, S.~Whitehead, A.~C. Berg, W.-Y. Lo, P.~Doll{\'a}r, and R.~Girshick.
\newblock \href{https://ai.meta.com/research/publications/segment-anything/}{Segment Anything}.
\newblock \emph{arXiv:2304.02643}, 2023.

\bibitem[Zhao et~al.(2023)Zhao, Ding, An, Du, Yu, Li, Tang, and Wang]{zhao2023fast}
X.~Zhao, W.~Ding, Y.~An, Y.~Du, T.~Yu, M.~Li, M.~Tang, and J.~Wang.
\newblock \href{https://arxiv.org/abs/2306.12156}{Fast segment anything}.
\newblock \emph{arXiv preprint arXiv:2306.12156}, 2023.

\bibitem[Oquab et~al.(2023)Oquab, Darcet, Moutakanni, Vo, Szafraniec, Khalidov, Fernandez, Haziza, Massa, El-Nouby, et~al.]{oquab2023dinov2}
M.~Oquab, T.~Darcet, T.~Moutakanni, H.~Vo, M.~Szafraniec, V.~Khalidov, P.~Fernandez, D.~Haziza, F.~Massa, A.~El-Nouby, et~al.
\newblock \href{https://arxiv.org/abs/2304.07193}{Dinov2: Learning robust visual features without supervision}.
\newblock \emph{arXiv preprint arXiv:2304.07193}, 2023.

\bibitem[Wen et~al.(2024)Wen, Yang, Kautz, and Birchfield]{wen2024foundationpose}
B.~Wen, W.~Yang, J.~Kautz, and S.~Birchfield.
\newblock \href{https://nvlabs.github.io/FoundationPose/}{Foundationpose: Unified 6d pose estimation and tracking of novel objects}.
\newblock In \emph{Proceedings of the IEEE/CVF Conference on Computer Vision and Pattern Recognition}, pages 17868--17879, 2024.

\end{thebibliography}

\newpage

%===============================================================================
\section{Appendix}
We provide additional information about our dataset in Appendix \ref{appendix:objects}, \ref{Appendix:data-gen}, and \ref{appendix:objects-data-quality}. Appendix \ref{appendix:exp-and-results} and \ref{appendix:real-world-experiments} show qualitative results and real-world experiment setup respectively.
\appendix

%===============================================================================
\section{Objects in the dataset}
\label{appendix:objects}
\begin{figure}[htbp]
    \begin{center}
            \includegraphics[width=\linewidth]{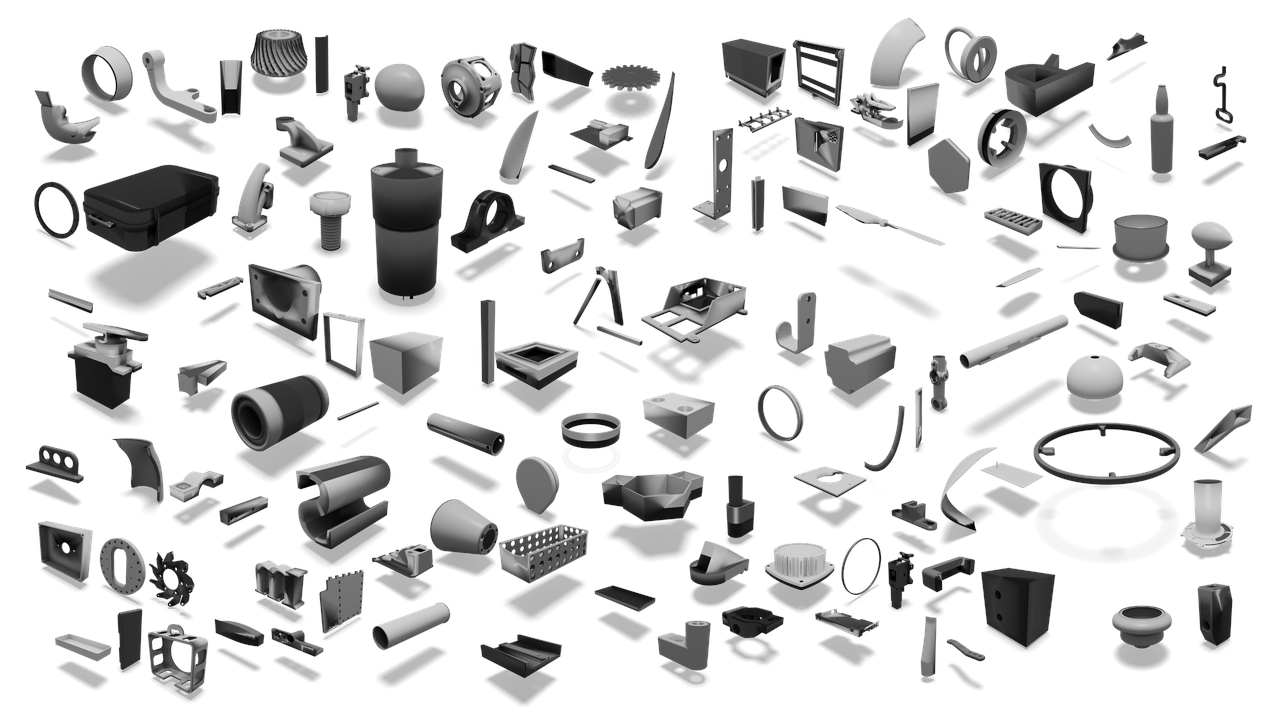}
            \captionof{figure}{A subset of the objects in the GraspFactory Dataset showing Geometric Variability.}
            \label{fig:objects-in-the-dataset}
        \end{center}
\end{figure}

Fig. \ref{fig:objects-in-the-dataset} shows a random subset of objects in the dataset. We compare the GraspFactory dataset with prior work in the literature and highlight the comparison in Table \ref{table:grasp_datasets}.

\begin{table*}[htbp]
\centering
\resizebox{\textwidth}{!}{ 
\begin{tabular}{c c c c c c c} 
\hline
\textbf{Dataset}  & \textbf{Planar/6D} & \textbf{Labels} & \textbf{Number of Objects} & \textbf{Candidate Grasps} & \textbf{Good Grasps/} & \textbf{Object-Centric } \\ 
& & & & & \textbf{Evaluated Grasps} & \textbf{Grasps} \\ \hline \hline
Cornell \cite{jiang2011cornell} & Planar & Manual & 240 & 8k & NA & \ding{55} \\ \hline
Jacquard \cite{depierre2018jacquard} & Planar & Sim & 11k & 1.1M & NA & \ding{55} \\ \hline
VMRD + Grasps \cite{zhang2019roi} & Planar & Manual & $\sim$15k & 100k & NA & \ding{55} \\ \hline
Columbia \cite{goldfeder2009columbia} & 6D & Analytical & 7256 & 238k & NA &  \checkmark\\ \hline
Dex-Net \cite{mahler2017dex} & 6D & Analytical & 1500 & 6.7M & NA & \ding{55} \\ \hline
6-DoF GraspNet \cite{asif2018graspnet} & 6D & Sim & 206 & 7.07M & NA & \ding{55} \\ \hline
GraspNet \cite{fang2020graspnet1b} & 6D & Analytical & 88 & 1.1B & NA & \ding{55} \\ \hline
EGAD \cite{morrison2020egad} & 6D & Analytical & 2331 & 233k & NA & \checkmark \\ \hline
ACRONYM \cite{eppner2021acronym} & 6D & Sim & 8872 & 17.7M & 10.5M/17.7M & \checkmark \\ \hline
GraspGen \cite{murali2025graspgen} & 6D & Sim & 8515 & 53.1M & NA & \checkmark \\
\hline
\textbf{GraspFactory (Ours)} & \textbf{6D} & \textbf{Sim} & \textbf{14,690} & \textbf{227.22M} & \textbf{12.2M/29.38M} & \checkmark  \\
\textbf{GraspFactory - Robotiq 2F-85 (Ours)} & \textbf{6D} & \textbf{Sim} & \textbf{33,710} & \textbf{391.38M} & \textbf{97.1M/164.16M} & \checkmark \\
\hline
\end{tabular}
}
\captionsetup{justification=centering}
\caption{Summary of various grasp datasets, highlighting labeling methods, number of objects and grasps. \\NA - no data available, \checkmark - dataset contains object centric grasps, \\\ding{55} - dataset contains grasps for a scene (images and point clouds).}
\label{table:grasp_datasets}
\end{table*}

%===============================================================================
\section{Data Generation}
\label{Appendix:data-gen}
\begin{figure}[htbp]
    \centering

    \begin{subfigure}[t]{0.49\linewidth}
        \centering
        \includegraphics[width=\linewidth]{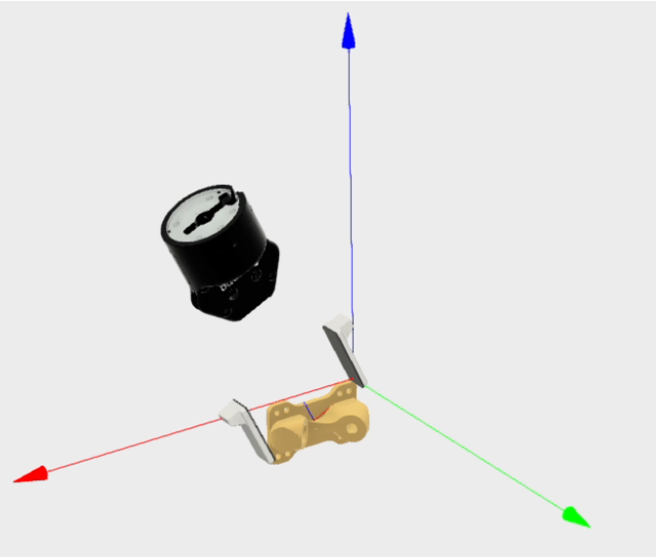}
        \caption{Collision check between an object and the gripper fingers using our internally developed robotics research software platform.}
        \label{fig:collision-check}
    \end{subfigure}
    \hfill
    \begin{subfigure}[t]{0.49\linewidth}
        \centering
        \includegraphics[width=\linewidth]{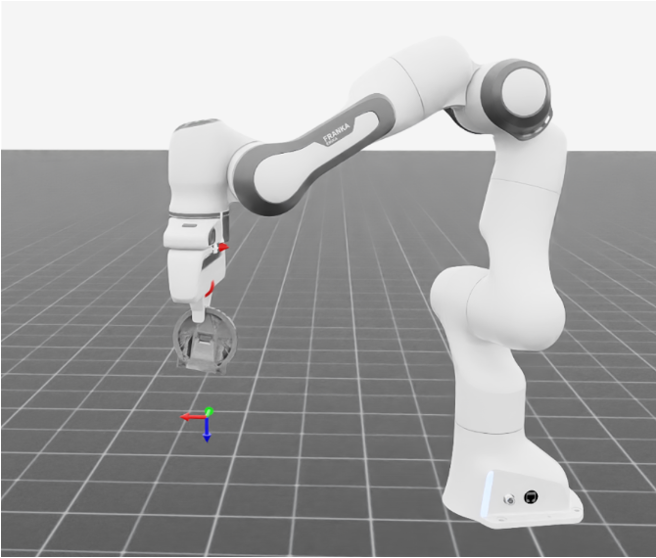}
        \caption{Single robot executing a sampled grasp.}
        \label{fig:eval-grasp-exec}
    \end{subfigure}

    \vspace{0.5em}

    \begin{subfigure}[t]{0.49\linewidth}
        \centering
        \includegraphics[width=\linewidth]{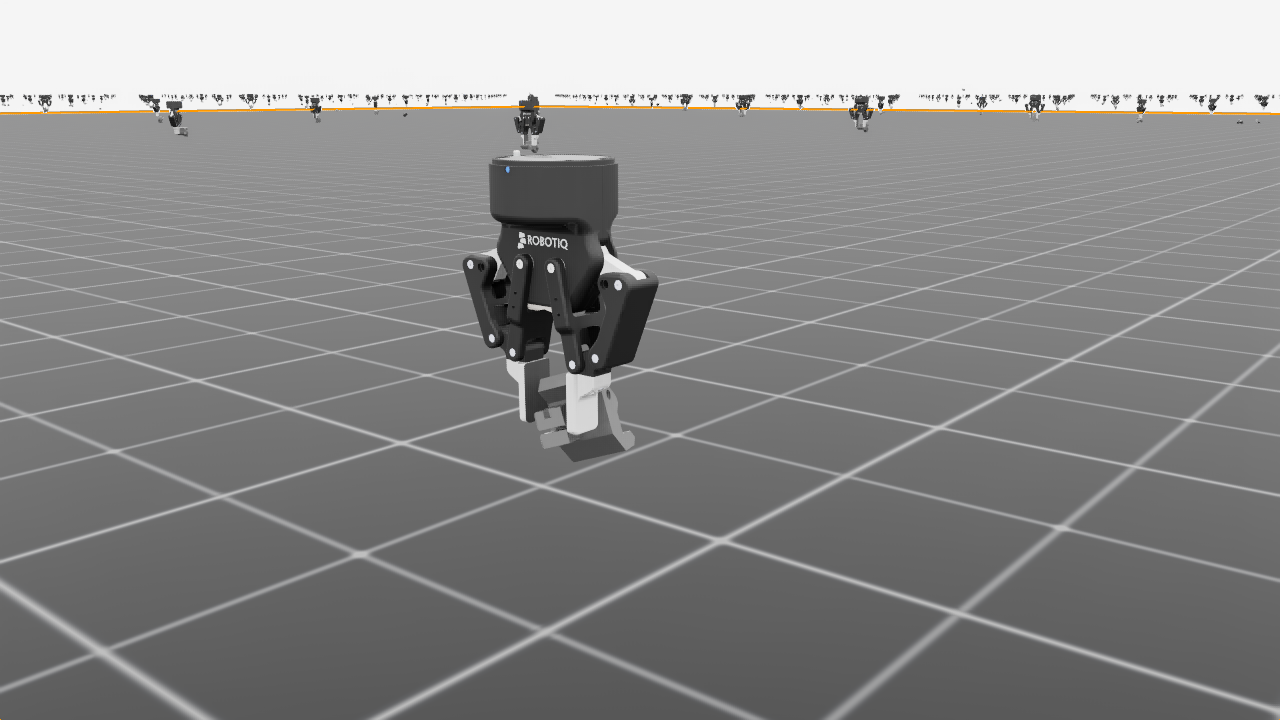}
        \caption{Isaac Sim evaluation environment with Robotiq 2F-85.}
        \label{fig:eval-isaac-sim-robotiq}
    \end{subfigure}
    \hfill
    \begin{subfigure}[t]{0.49\linewidth}
        \centering
        \includegraphics[width=\linewidth]{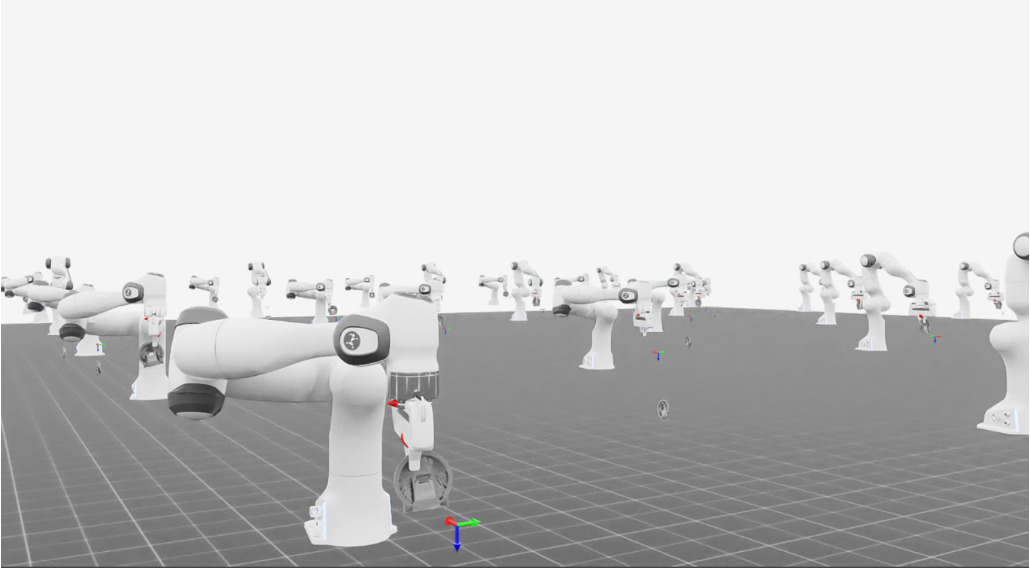}
        \caption{Isaac Sim evaluation environment with Franka Panda Robot with Panda Hand.}
        \label{fig:eval-isaac-sim}
    \end{subfigure}

    \captionsetup{justification=centering}
    \caption{Simulation environments for evaluating collision checks, sampled grasp execution, and physics-based evaluation.}
    \label{fig:isaac-sim-env}
\end{figure}

We use an internally developed robotics research platform to perform collision check between the gripper fingers and the sampled objects as shown in Fig. \ref{fig:collision-check}. Simulated Robotiq 2F-85 is shown in Fig. \ref{fig:eval-isaac-sim-robotiq} and Franka Panda Hand in Isaac Sim to test physical feasibility is shown in Fig. \ref{fig:eval-isaac-sim}.

\begin{figure}[ht]
    \centering
    \includegraphics[width=0.6\linewidth, keepaspectratio]{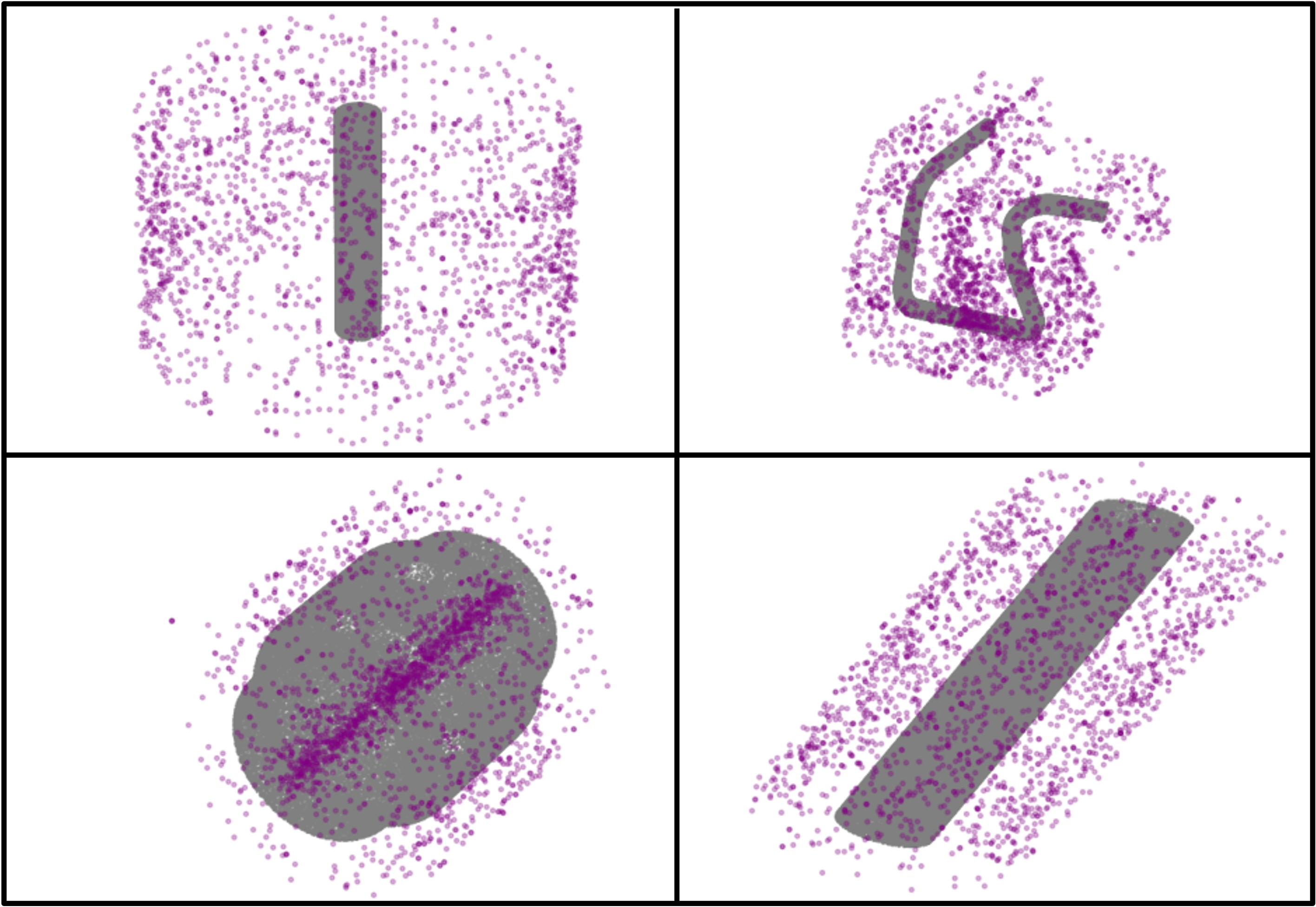}
    \captionsetup{justification=centering}
    \caption{Qualitative analysis of grasp coverage for four randomly selected objects from the dataset. Purple points represent gripper positions around the objects.}
    \label{fig:grasp-coverage}
\end{figure}

The evaluation pipeline for ABC-Grasp dataset generation in simulation assumes uniform physical properties, including mass and the coefficient of friction, across all objects in the dataset to ensure computational feasibility. We note that variations in mass and friction may influence grasp stability and robustness for objects with slippery or uneven surfaces. Additionally, we treat all objects in the dataset as rigid.

%===============================================================================
\section{Data Quality}
\label{appendix:objects-data-quality}
A plot of the location of successful grasps, shown in Fig. \ref{fig:grasp-coverage} shows that our method covers the entire space around the objects.

\begin{figure*}[t]
    \centering
    \includegraphics[width=0.8\linewidth]{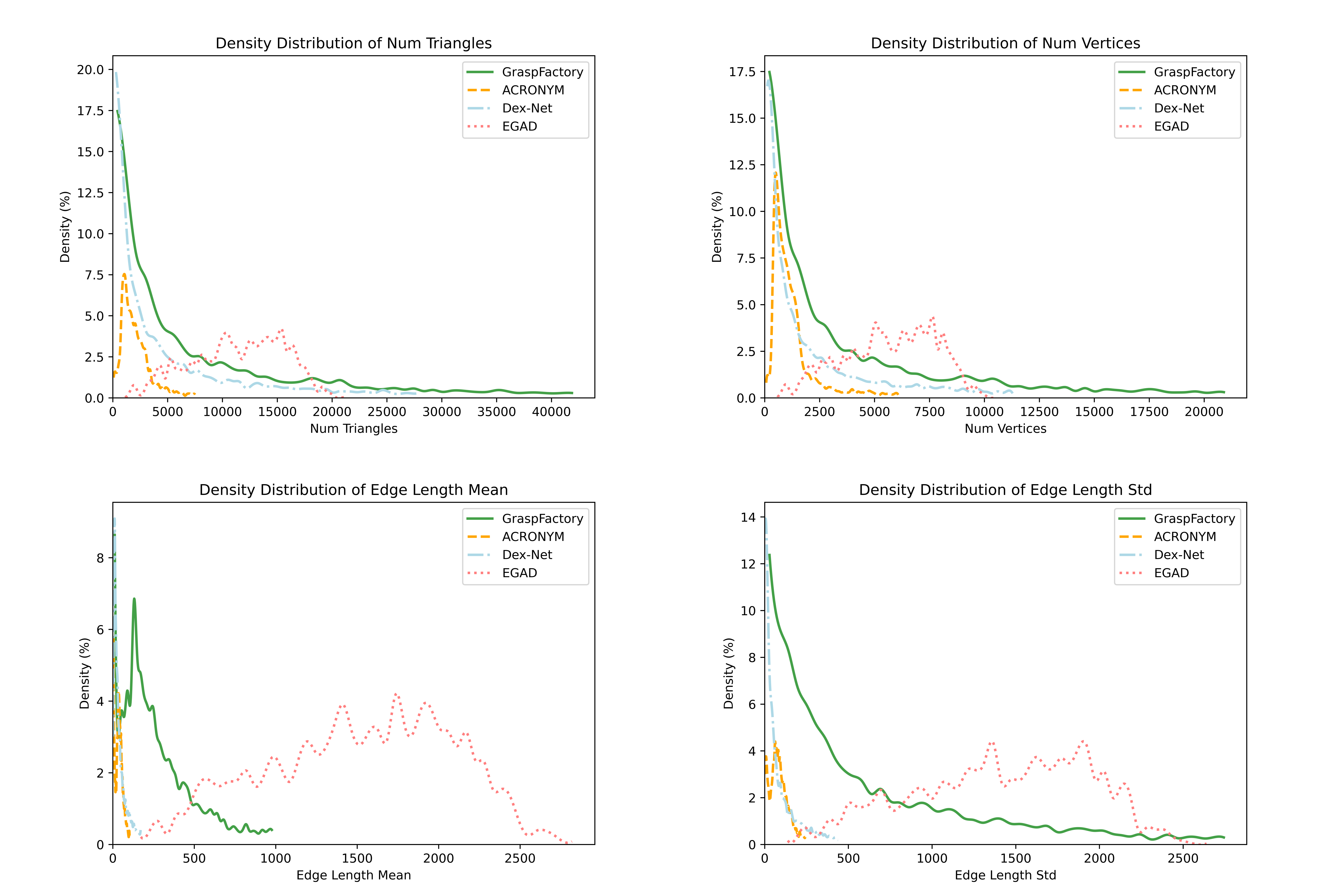}
    \captionsetup{justification=centering}
    \caption{Density distribution curves for the number of triangles, number of vertices and edge lengths' mean and standard deviation for
    GraspFactory (Ours), ACRONYM \cite{eppner2021acronym}, Dex-Net \cite{mahler2017dex}, EGAD \cite{morrison2020egad} showing a larger variance in GraspFactory. \\
    EGAD shows more uniform spread in Edge Length metrics, but contains approximately seven times fewer objects and is less representative of real-world objects.}
    \label{fig:histograms}
\end{figure*}

Fig. \ref{fig:histograms} presents metrics such as the \textit{number of triangles}, \textit{number of vertices}, and \textit{edge length statistics} for the GraspFactory, ACRONYM, Dex-Net, and EGAD datasets.  The graphs demonstrate that GraspFactory exhibits a wider spread compared to both ACRONYM and Dex-Net. While EGAD shows a more uniform distribution than GraspFactory, GraspFactory contains approximately seven times more objects, and its objects better align with those encountered in the real world compared to the EGAD dataset.

These metrics were chosen to highlight geometric diversity as they are directly related to the structural complexity of the meshes, serving as quantifiable indicators of geometric diversity. They are also computationally efficient to calculate and provide an immediate sense of the detail in a CAD model.

%===============================================================================
\section{Qualitative results}
\label{appendix:exp-and-results}

\begin{figure}[ht]
    \centering
    \includegraphics[width=\linewidth]{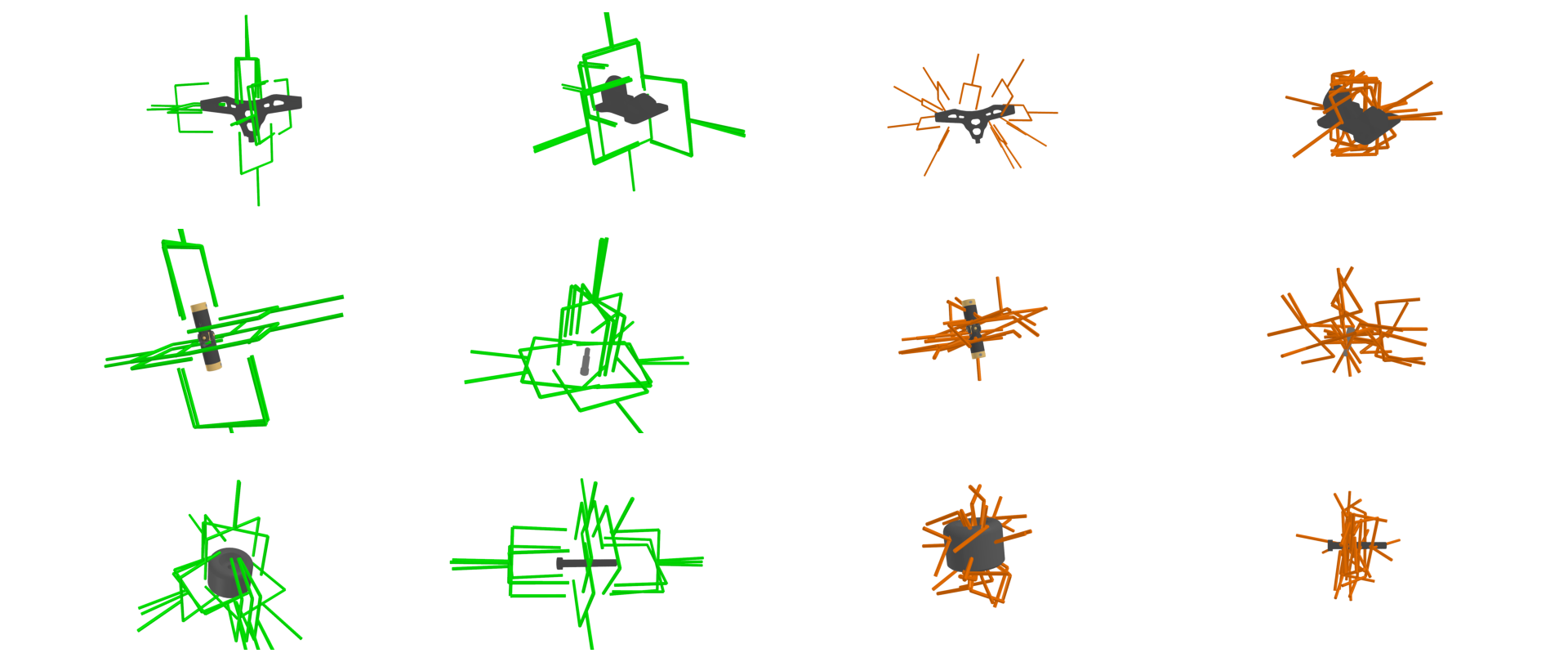}
    \captionsetup{justification=centering}
    \caption{Qualitative comparison of grasps generated for unseen objects by the model trained  on GraspFactory (ours, left two columns in green) and the model trained on ACRONYM (right two columns in orange).}
    \label{fig:grasps-abc-acronym}
\end{figure}

We show the qualitative results of the model trained on GraspFactory and ACRONYM datasets in Fig. \ref{fig:grasps-abc-acronym}. The model trained on ACRONYM produces grasps that intersect with the objects, whereas, the model trained on our GraspFactory dataset produces grasps uniformly around the objects.

%===============================================================================
\section{Real-World Experiments}
\label{appendix:real-world-experiments}

\begin{figure}[htbp]
    \centering
    \includegraphics[width=\columnwidth, height=0.7\textheight, keepaspectratio]{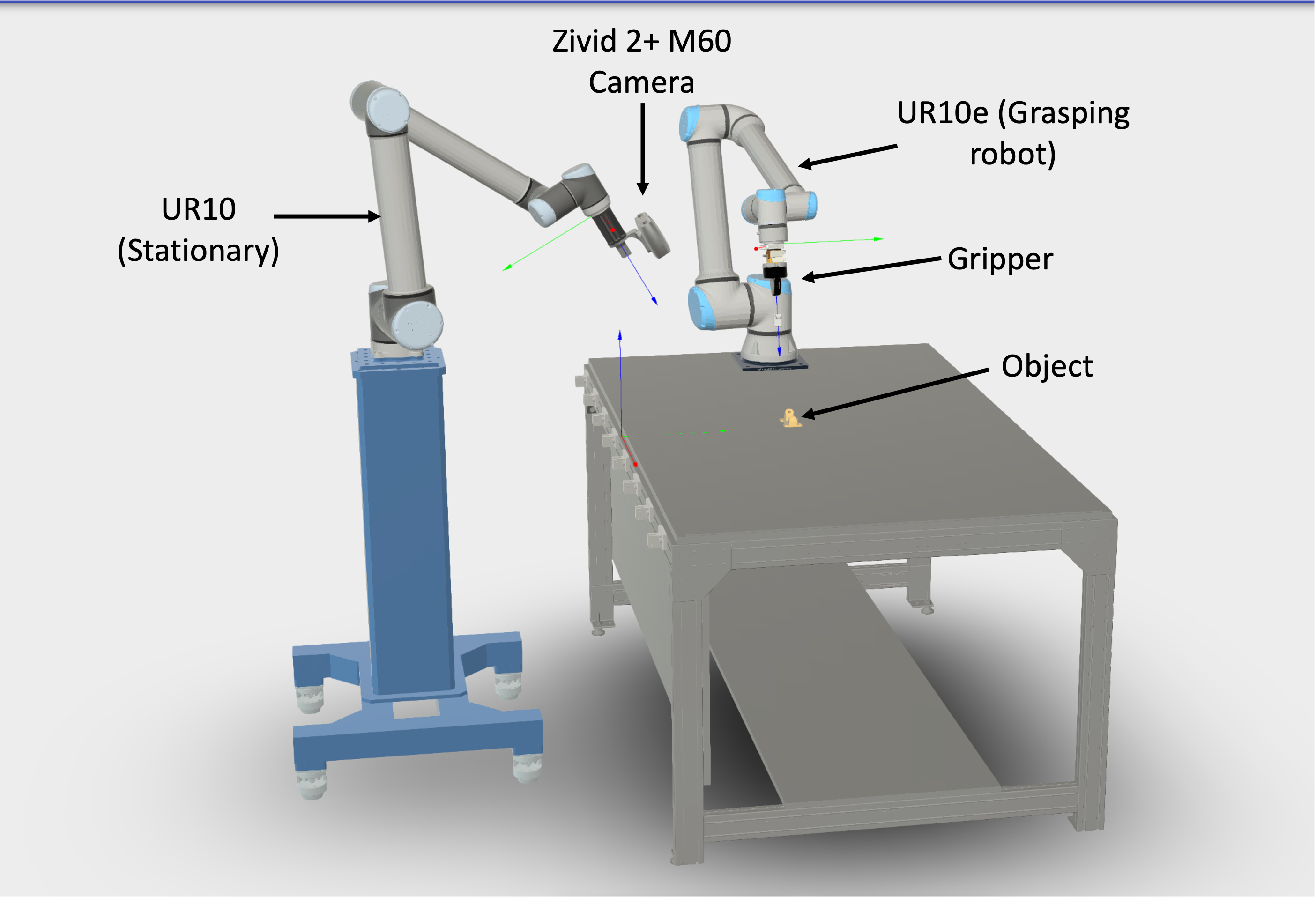}
    \caption{Workcell setup for real-world experiments. We use a UR-10e robot equipped with a Robotiq 2F-85 gripper for grasping. Zivid 2+ M60 camera is mounted on a UR-10 robot.}
    \label{fig:workcell}
\end{figure}

\begin{figure}[htbp]
    \centering
    \includegraphics[width=\columnwidth, height=0.9\textheight, keepaspectratio]{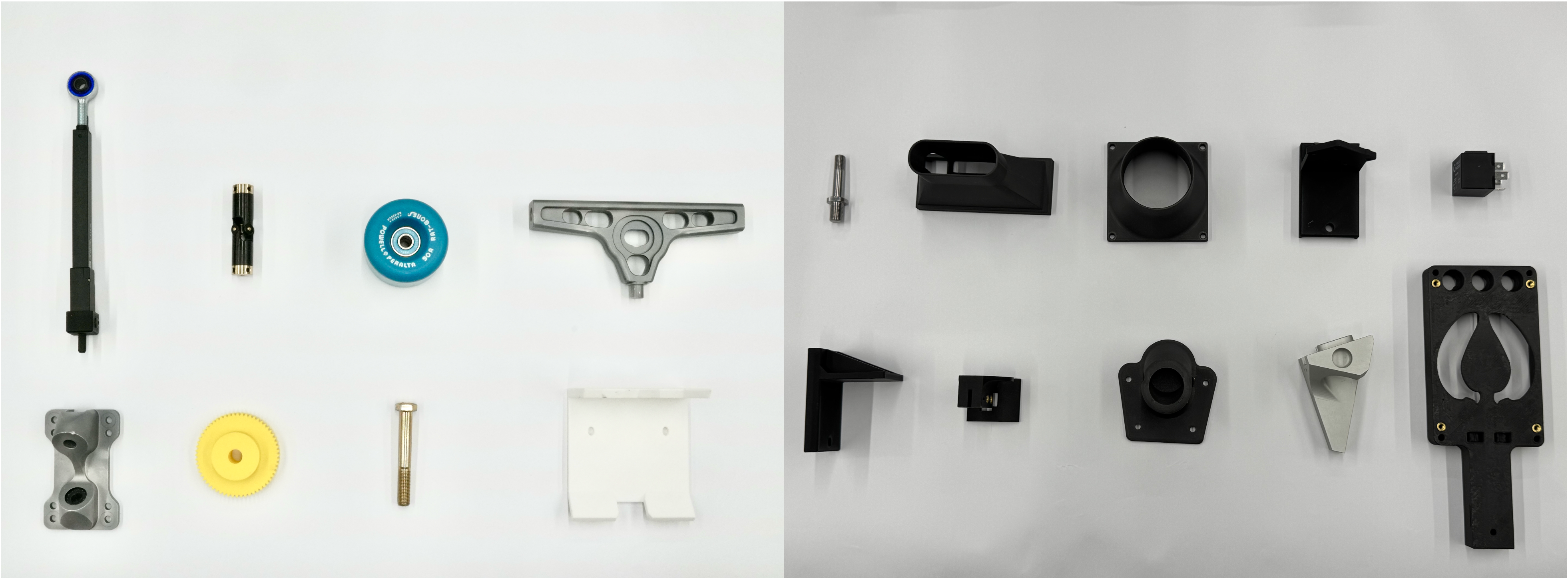}
    \caption{Objects used for real-world experiments}
    \label{fig:real-objects}
\end{figure}

We perform physical experiments on 18 real-world objects shown in Fig. \ref{fig:real-objects}. The workcell setup for the experiment consists of a Zivid 2+ M60 camera mounted on a UR-10 robot, and grasping is performed by a UR-10e robot, equipped with a Robotiq 2F-85 two-fingered gripper \footnote{Due to an unanticipated lack of availability of our Franka Panda robot, we chose to use a UR10e robot and Robotiq 2F-85 gripper for testing on real-hardware.} as shown in Fig. \ref{fig:workcell}. We note that the model was trained on grasps that were validated using a Franka Panda robot with Franka hand in simulation. This gripper has a finger width of \(18mm\), while our real-world evaluation is performed using a Robotiq 2F-85 two-fingered gripper whose finger width is \(22mm\).

\begin{figure}[ht]
\centering
\includegraphics[width=0.6\columnwidth]{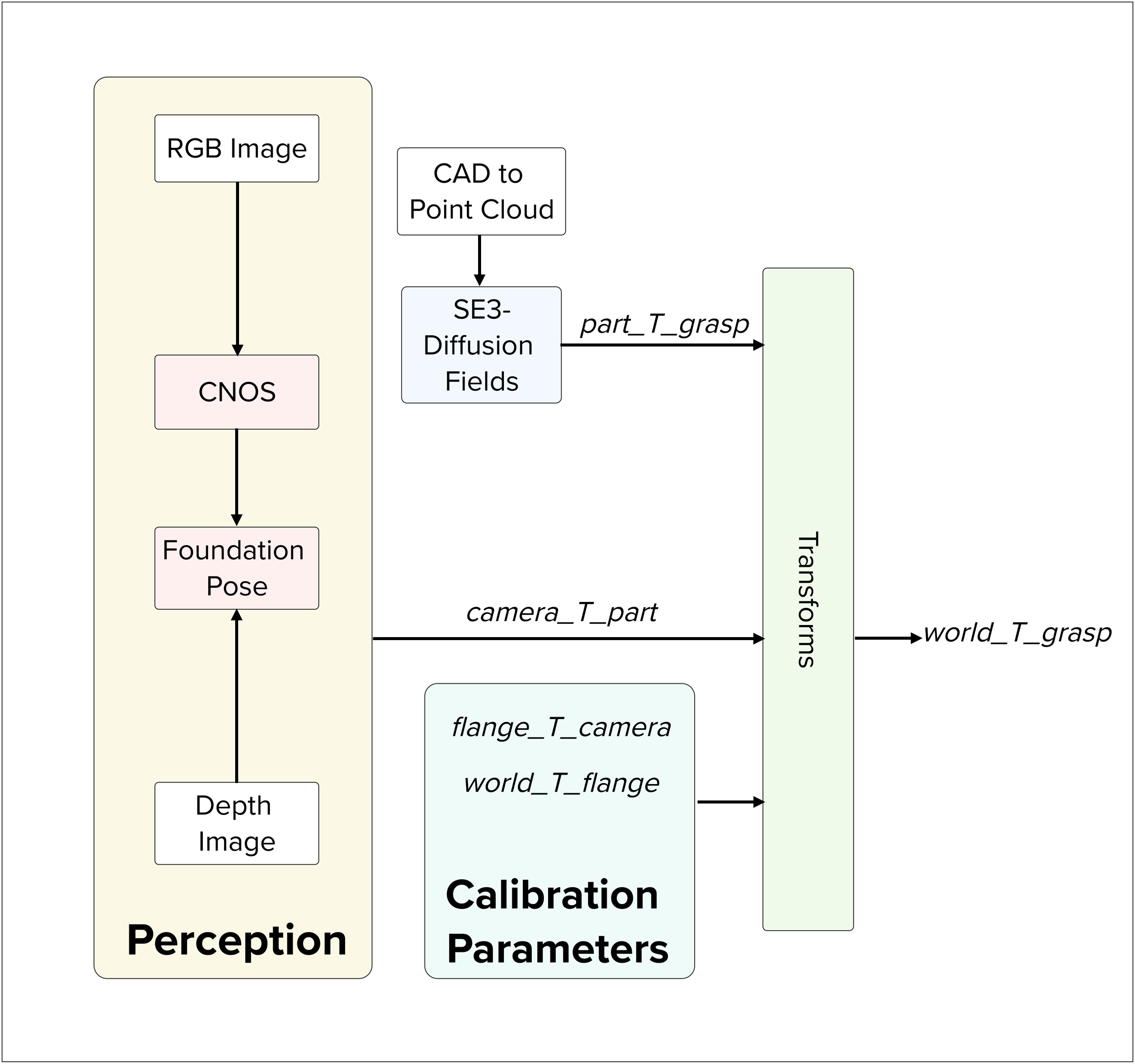}
\captionsetup{justification=centering}
\caption{Pipeline for real-world experiments}
\label{fig:real-world-pipeline}
\end{figure}

\begin{figure}[htbp]
    \includegraphics[width=\linewidth, keepaspectratio]{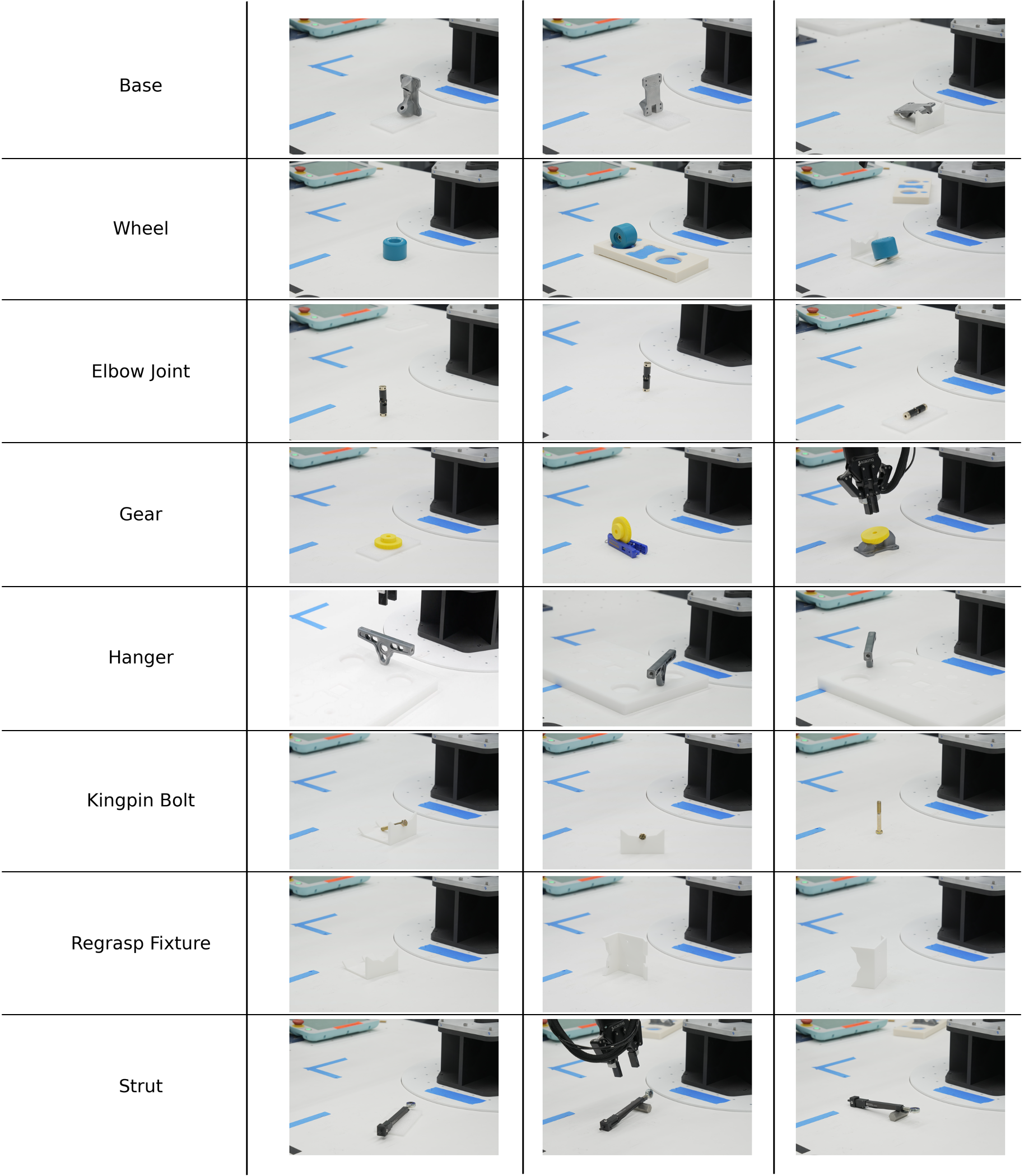}
    \captionsetup{justification=centering}
    \caption{Real world experiment with eight objects, each in three random stable poses.}
    \label{fig:real-world-experiment-poses}
\end{figure}

\begin{figure}[htbp]
    \centering
    \includegraphics[width=\linewidth]{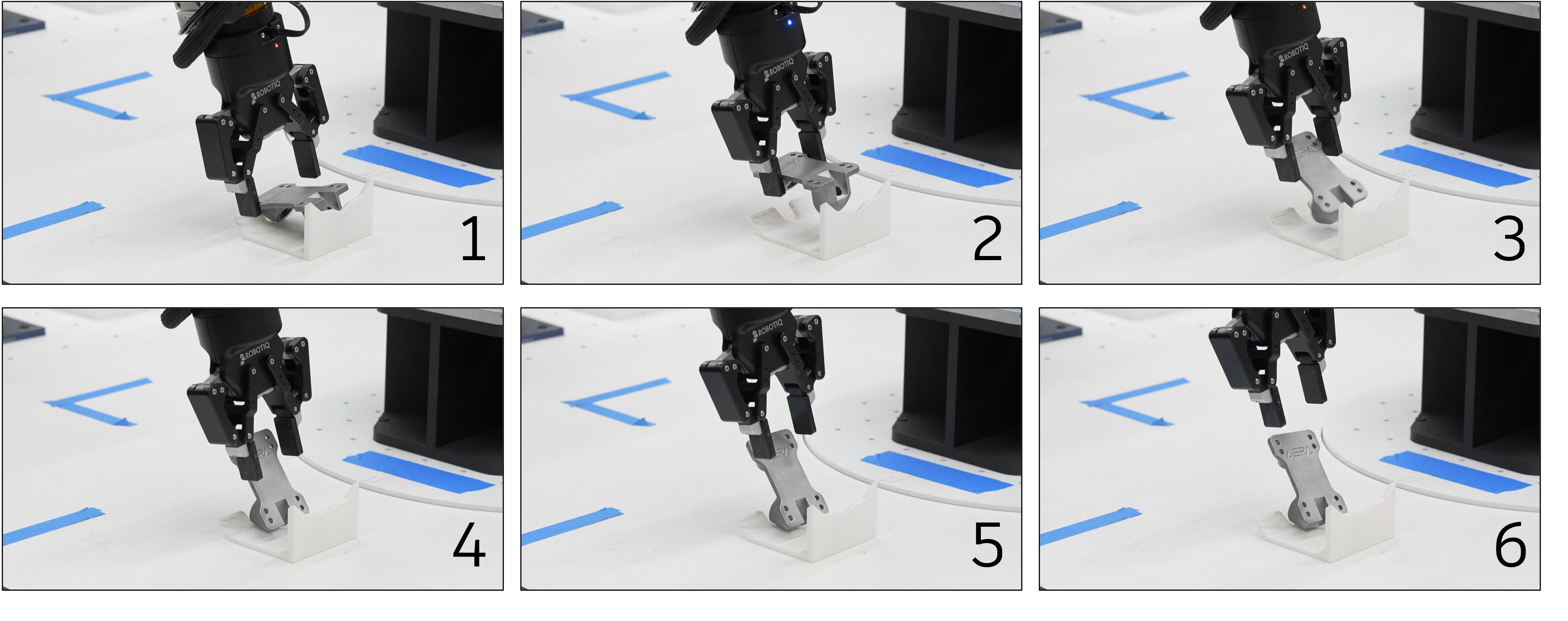}
    \caption{Sequence of frames depicting the \textit{Base} slipping out of the fingers. Frame sequence numbers are embedded in the images.}
    \label{fig:failure-sequence}
\end{figure}

In each experiment, we first place individual objects in front of the robot and implement a perception based pipeline, as shown in Fig. \ref{fig:real-world-pipeline}, to locate the object with respect to the robot. Our perception pipeline builds on  CNOS presented by \citet{nguyen2023cnos}, a method that utilizes segmentation proposals generated from the captured RGB image using the Segment Anything Model \cite{kirillov2023segany} or Fast Segment Anything model (FastSAM) \cite{zhao2023fast} to localize an object of interest in the scene. CNOS matches the DINOv2 cls \cite{oquab2023dinov2} tokens of the proposed segmentation regions against tokens of object templates that are pre-rendered using their CAD models. We use this localization information to segment the point cloud of the object captured by the camera.

With the object localized in the scene, we use the model-based setup of FoundationPose \cite{wen2024foundationpose}, which uses the CAD model and the segmented point cloud of the object (which we obtain from CNOS) to estimate the object's 6-DoF pose in the camera frame, denoted by $_c{\textit{T}}_o$. We then use the camera extrinsics and robot calibration parameters to transform the computed grasps into the world-coordinate frame, as described by the equation below:

\begin{equation}
\label{eq:grasp-transformation}
{}_{w}T_{g} = {}_{w}T_{r} \cdot {}_{r}T_{f} \cdot {}_{f}T_{c} \cdot {}_{c}T_{o} \cdot {}_{o}T_{g}
\end{equation}

where, \textit{T} is a transformation matrix defined as shown in Eq. \ref{eq:transformation_matrix}, \textit{w} is the world frame, \textit{r} is the robot frame, \textit{f} is the robot flange frame, \textit{c} is the camera frame, \textit{o} is the part or the object frame.

We choose 18 diverse set of objects, shown in Fig. \ref{fig:real-objects} to evaluate the model trained on our dataset in real-world settings.

Fig. \ref{fig:failure-sequence} shows that the flat finger geometry leads to unstable grasps near the center of mass of the \textit{Base}, a behavior also observed in our simulation experiments. Since the model we train does not account for finger geometry, incorporating this factor could help ensure the generation of only stable grasps.

%===============================================================================
\end{document}